%% file: paper.tex
\documentclass[letterpaper]{article} 
\usepackage{aaai25}  
\usepackage{times}  
\usepackage{helvet}  
\usepackage{courier}  
\usepackage[hyphens]{url}  
\usepackage{graphicx} 

\usepackage{booktabs}
\usepackage{mathtools}
\usepackage{amsthm}
\usepackage{multirow}
\usepackage{subcaption} 
\usepackage[capitalize]{cleveref}

\usepackage{natbib}  
\usepackage{caption} 
\usepackage{algorithm}
\usepackage{algorithmic}

\usepackage{newfloat}
\usepackage{listings}
\DeclareCaptionStyle{ruled}{labelfont=normalfont,labelsep=colon,strut=off} 
\floatstyle{ruled}
\newfloat{listing}{tb}{lst}{}
\floatname{listing}{Listing}
\usepackage{color} 

\usepackage{amsfonts}
\usepackage{amsmath}

\def\q{\mathbf{q}}
\def\R{\mathbb{R}}

\def\b{\mathbf{b}}

\def\r{\mathbf{r}}

\def\u{\mathbf{u}}

\def\m{\mathbf{m}}

\def\g{\mathbf{g}}
\def\x{\mathbf{x}}

\def\v{\mathbf{v}}

\def\q{\mathbf{q}}
\def\w{\mathbf{w}}

\def\R{\mathbb{R}}

\title{BEE: Metric-Adapted Explanations via Baseline Exploration-Exploitation}
\author{
    Oren Barkan\textsuperscript{\rm 1}\equalcontrib,
    Yehonatan Elisha\textsuperscript{\rm 2}\equalcontrib,
    Jonathan Weill\textsuperscript{\rm 2}, Noam Koenigstein\textsuperscript{\rm 2}
}
\affiliations{
    \textsuperscript{\rm 1}The Open University, Israel\\
    \textsuperscript{\rm 2}Tel Aviv University, Israel

}

\usepackage{bibentry}

\begin{document}

\maketitle

\begin{abstract}
Two prominent challenges in explainability research involve 1) the nuanced evaluation of explanations and 2) the modeling of missing information through baseline representations. The existing literature introduces diverse evaluation metrics, each scrutinizing the quality of explanations through distinct lenses. Additionally, various baseline representations have been proposed, each modeling the notion of missingness differently. Yet, a consensus on the ultimate evaluation metric and baseline representation remains elusive. This work acknowledges the diversity in explanation metrics and baselines, demonstrating that different metrics exhibit preferences for distinct explanation maps resulting from the utilization of different baseline representations and distributions. To address the diversity in metrics and accommodate the variety of baseline representations in a unified manner, we propose Baseline Exploration-Exploitation (BEE) - a path-integration method that introduces randomness to the integration process by modeling the baseline as a learned random tensor. This tensor follows a learned mixture of baseline distributions optimized through a contextual exploration-exploitation procedure to enhance performance on the specific metric of interest. By resampling the baseline from the learned distribution, BEE generates a comprehensive set of explanation maps, facilitating the selection of the best-performing explanation map in this broad set for the given metric. Extensive evaluations across various model architectures showcase the superior performance of BEE in comparison to state-of-the-art explanation methods on a variety of objective evaluation metrics.
Our code is available at: \url{https://github.com/yonisGit/BEE}
\end{abstract}


\input{latex/1_intro}

\input{latex/2_related}
\input{latex/3_method}
\input{latex/4_experiments}
\input{latex/5_conclusion}

\bibliography{aaai25}
\appendix
\input{latex/6_appendix}

\end{document}

%% file: latex/1_intro.tex
\section{Introduction}
\label{sec:intro}
Deep learning models have demonstrated remarkable success across a spectrum of tasks in computer vision~\cite{he2016resnet,dosovitskiy2020image, carion2020end}, natural language processing~\cite{vaswani2017attention, devlin2018bert, barkan2020bayesian, barkan2020scalable, barkan2021representation, touvron2023llama}, recommender systems~\cite{he2017neural, barkan2016item2vec, barkan2020attentive, barkan2021anchor, barkan2021cold, katz2022learning}, and audio processing~\cite{barkan2019inversynth, barkan2023inversynth, engel2020ddsp, kong2020diffwave}. Despite their accomplishments, these models frequently function as opaque systems, introducing challenges in comprehending their predictions.
Consequently, the field of Explainable AI (XAI) has emerged, dedicated to developing methods that illuminate the decision rationale of machine learning models across diverse application domains~\cite{simonyan2013deep, malkiel2022interpreting, gaiger2023not,barkan2020explainable, barkan2023learning, barkan2024counterfactual, barkan2024learning,barkan2024llm,barkan2024improving}. In the context of computer vision, XAI techniques aim to generate explanation maps highlighting input regions responsible for the model's predictions~\cite{selvaraju2017grad, chefer2021transformer}. 
For example, Integrated Gradients (IG)~\cite{SundararajanTY17}, which produces explanation maps by integrating gradients along a linear path between the input image and a baseline representation (acting as a reference representing missing information). Nevertheless, a challenge persists in selecting the appropriate baseline, as different types of baselines model missingness differently, resulting in variations in the explanation maps. Despite the exploration of various baselines in the literature, no consensus has emerged on the ultimate baseline representation~\cite{ancona2017towards, fong2017interpretable, sturmfels2020visualizing}. Another prominent challenge in XAI revolves around evaluating the effectiveness of the generated explanations. Various evaluation metrics have been proposed in the literature~\cite{chefer2021transformer,petsiuk2018rise,kapishnikov2019xrai,chattopadhay2018grad}, each assessing the quality of explanation maps from different perspectives. As a result, distinct explanation metrics may promote different explanation maps, and currently, there is no universally agreed-upon evaluation metric for assessing the goodness of explanations. Acknowledging the diverse landscape of evaluation metrics and baseline representations, this paper introduces Baseline Exploration-Exploitation (BEE) - a path-integration method utilizing an exploration-exploitation (EE) mechanism to adapt the baseline distribution (and hence the resulting explanation) w.r.t. the specific metric of interest. This approach offers an effective way to address the diversity in metrics and baselines in a cohesive manner. BEE integrates on the intermediate representations (and their gradients) produced by different network layers, thereby generating explanation maps at multiple levels of abstractions and various scales. The key innovation of BEE lies in introducing randomness to the integration process by modeling the baseline representation as a random tensor. This tensor adheres to a mixture of baseline distributions learned through an offline pretraining phase. The pretraining phase, employs contextual EE of the baseline distribution to optimize performance on the specific metric at hand. Given a test instance (context), BEE generates a pool of candidate explanation maps produced from a corresponding set of baselines sampled from the pretrained baseline distribution in parallel. Subsequently, the explanation map that performs the best on the metric is selected. For further enhancement, BEE can continually employ contextual EE on the specific instance during inference, facilitating instance-specific finetuning of the pretrained baseline distribution w.r.t. the metric of interest. The effectiveness of BEE is demonstrated through extensive evaluation on various model architectures. The results showcase that the pretrained BEE consistently outperforms latest state-of-the-art explanation methods across all objective evaluation metrics. Furthermore, when inference-time finetuning is permitted, the finetuned BEE yields additional performance gains over the pretrained BEE, affirming its viability as a superior option.

Our contributions are summarized as follows: 1) We highlight and demonstrate the challenges arising from the absence of a universally agreed-upon evaluation metrics and baseline representations in XAI: Different metrics promote different explanation maps resulting from different types of baselines. 2) To address these challenges, we introduce BEE - a novel contextual EE-based path-integration method. BEE models the baseline representation as a random tensor sampled from a mixture of distributions, accommodating various types of baselines, and facilitating adaptive explanations w.r.t. the metric at hand. 3) Through extensive evaluation against 13 explanation methods, both the pretrained and finetuned BEE versions emerge as a new state-of-the-art in XAI, outperforming latest state-of-the-art methods across 8 evaluation metrics and 5 CNN and ViT architectures.

%% file: latex/2_related.tex
\section{Related Work}
\label{sec:related}
Literature on explanation methods for CNNs has
grown with several broad categories
of approaches: perturbation-based methods~\cite{fong2017interpretable, barkan2023learning}, gradient-free methods~\cite{zhou2018interpreting,zhou2016learning,zeiler2014visualizing,Wang2020ScoreCAMSV}, and gradient-based methods~\cite{selvaraju2017grad,Srinivas2019FullGradientRF},
which include path-integration methods~\cite{SundararajanTY17, xu2020attribution,Barkan_2023_ICCV,barkan2023six,barkan2023deep,elisha2024probabilistic}.
The most relevant line of work to this paper are path-integration methods~\cite{SundararajanTY17, xu2020attribution, Barkan_2023_ICCV}. IG~\cite{SundararajanTY17} integrates over the interpolated image gradients. Blur IG~\cite{xu2020attribution} integrates gradients over a path that progressively removes Gaussian blur from the attributed image. Our method, provides two significant differences w.r.t. the aforementioned works: First, BEE extends the integrand beyond the gradient itself, incorporating information from both the internal network representations and their gradients. This capability facilitates the generation of explanation maps at multiple levels of abstractions and resolution and has proven effective~\cite{Barkan_2023_ICCV}. Second, through the introduction of an EE procedure, BEE can learn baseline distributions customized for the specific metric of interest. This characteristic enables BEE to draw a set of explanation maps resulting from the sampled baselines and select the best-performing one w.r.t. the given metric. Early explanation methods for transformer explainability involved leveraging the inherent attention scores to gain insights into the input~\cite{carion2020end}. However, a challenge arises in combining scores from different layers. Simple averaging of attention scores for each token, for instance, tends to blur the signal~\cite{abnar2020quantifying}. 
Chefer et al.\cite{chefer2021transformer} introduced Transformer Attribution (T-Attr), a class-specific Deep Taylor Decomposition method in which relevance propagation is applied for positive and negative attributions. As a follow-up work, the Generic Attention Explainability (GAE)~\cite{chefer2021generic} was introduced as a generalization of T-Attr for explaining Bi-Modal transformers. In contrast to T-Attr and GAE, which are specifically designed for transformers, BEE is a versatile approach that can generate explanations for both CNN and ViT models in a unified manner. More recently, Iterated Integrated Attributions (IIA)~\cite{Barkan_2023_ICCV} proposed a generalization of IG through an iterated integral. While IIA uses a constant baseline for the integration process, BEE models the baseline as a random tensor, enabling the generation of multiple distinct explanation maps. The stochastic nature of BEE allows for the selection of the optimal explanation map from this diverse set, tailored to the specified metric.


%% file: latex/3_method.tex
\section{Baseline Exploration-Exploitation}
Let $\x \in \R^{d_0}$ be the input image. Let $f:\R^{d_0} \rightarrow \R^{d_L}$ be a neural network consisting of $L$ layers, each producing a representation $\x^l \in \R^{d_l}$, and $\x^0:=\x$. The final layer produces the prediction $f(\x)$, in which the score for the class $y$ is given by $f_y(\x)$. Our goal is to produce an explanation map $\m\in\R^{d_0}$ w.r.t. the class $y$, in which each element $\m_i$ represents the attribution of the prediction $f_y(\x)$ to the element $\x_i$. IG~\cite{SundararajanTY17} enables the creation of an explanation map by defining a linear path between a baseline representation $\b$ and $\x$ via the parameterization:
\begin{equation}
\label{eq:v-ig}
\v=(1-a)\b+a\x\,\, \text{with} \,\, a\in[0,1],
\end{equation}
and accumulating the gradients along this path as follows:

\begin{equation}
\begin{split} 
\label{eq:ig}
\m_{IG}&=\int_0^1 \frac{\partial f_y(\v)}{\partial \v} \circ \frac{\partial \v}{\partial a}da\approx \frac{\x-\b}{n} \circ \sum_{k=1}^n \frac{\partial f_y(\v)}{\partial \v},
\end{split}
\end{equation}
where $\circ$ denotes the elementwise product, and the approximation is obtained by setting $a=\frac{k}{n}$ in Eq.~\ref{eq:v-ig}.
A path between $\b$ and $\x$ symbolizes the transition from the uninformative baseline $\b$, essentially representing missing information, to the informative image $\x$. Therefore, it is crucial to design the baseline representation such that it aligns with the concept of missing information. In Sec.~\ref{sec:abs}, we present the types of baselines considered in this work, and the BEE procedure that enables adaptive sampling of baselines.
BEE constructs an explanation map by integrating functions that incorporate information from both the internal network representations and their gradients. 
A distinctive characteristic of BEE is its utilization of stochastic integration paths, wherein the baseline $\b$ is sampled from a distribution $\mathcal{D}$ learned through a contextual EE procedure. This procedure facilitates adaptive baseline sampling from different types of baseline distributions, resulting in the generation of diverse explanation maps. Given the challenge of quantifying the quality of an explanation map, and considering the absence of a universally agreed-upon metric, BEE enables the selection of the best-performing explanation map (or their combination / aggregation) from a diverse set of explanations w.r.t. the specific metric of interest. We start by describing the BEE explanation map generation process, assuming the baseline distribution $\mathcal{D}$ is \textbf{given} (Sec.~\ref{sec:explanation-map-construction}). Then, in Sec.~\ref{sec:abs}, we introduce the BEE procedure that optimizes the baseline distribution $\mathcal{D}$ per metric.
\subsection{The BEE explanation map construction}
\label{sec:explanation-map-construction}
Let $f^l:\R^{d_l} \rightarrow \R^{d_L}$ be a sub-network of $f$ taking an input $\x^l$ and producing the final prediction $f^l_y(\x^l)$. Given a prescribed number of trials $T$, BEE samples $T$ baselines $\{\b^{lt}\}_{t=1}^T$ from the baseline distribution $\mathcal{D}$ and computes a set of explanation maps $M^l=\{\m^{lt}\}_{t=1}^T$ as follows:
\begin{equation}
\label{eq:abs-map}
\m^{lt}= u\left(\frac{\x^l-\b^{lt}}{n} \circ \sum_{k=1}^n \psi\left(\frac{\partial f^l_y(\v^{lt})}{\partial \v^{lt}}, \v^{lt}\right)\right),
\end{equation}
where $\v^{lt}=(1-\frac{k}{n})\b^{lt} + \frac{k}{n}\x^l$ is the interpolated representation, $\psi$ is a function combining information from $\v^{lt}$ and its gradients, and $u$ is a function transforming the resulting explanation map to match the original spatial dimensions of $\x$.
The stochastic nature of BEE enables the formation of multiple explanation maps $M^l$. By considering explanation maps obtained from different network layers, we form a superset of explanation maps $M^{\mathcal{I}}=\cup_{l\in\mathcal{I}} M^l$, where $\mathcal{I}$ is a set of selected layer indexes. Finally, given a metric of interest $s$ that provides an assessment score $s(\m)$ for the goodness of the explanation map $\m$, the BEE explanation map is defined by:
\begin{equation}
\label{eq:abs-map-final}
\m_{\text{BEE}}= \underset{\m \in M^{\mathcal{I}}}{\operatorname{argmax}} \,s(\m).
\end{equation}
Therefore, $\m_{\text{BEE}}$ is the explanation map that performs the best on the metric $s$ among the maps in $M^{\mathcal{I}}$. It is important to clarify that BEE selects the best-performing explanation from a relatively small set of candidate explanation maps. This set is not guaranteed to include the optimal explanation map.
\subsection{Learning the baseline distribution with BEE}
\label{sec:abs}
In this section, we describe the BEE procedure that produces the (adpative) baseline distribution $\mathcal{D}$ (which was assumed to be given in Sec.~\ref{sec:explanation-map-construction}).
Previous works have extensively discussed and compared various notions of missingness in the context of attribution baselines~\cite{fong2017interpretable,ancona2017towards,kindermans2019reliability,sturmfels2020visualizing}. However, the challenge remains in selecting the appropriate baseline.
In this work, we explore different types of baselines: 1) The constant baseline (\textbf{Constant}): In~\cite{SundararajanTY17}, the authors employed the black baseline. Generally, a value can be drawn from a valid range to form a constant baseline based on this value. Since BEE integrates on the intermediate representation layers in the network (activations), the valid ranges are set based on the minimum and maximum value in each channel of the activation. However, this baseline may not accurately model missingness. For instance, using a constant black image as a baseline in IG may not highlight black pixels as important, even if these pixels constitute the object of interest. 2) The blur baseline (\textbf{Blur}): In~\cite{fong2017interpretable}, the authors used a blurred version of the image (or activation in the case of BEE) as a domain-specific technique to represent missing information. This approach is appealing due to its intuitive capture of the concept of missingness in images. The blur operation is controlled by a parameter that determines the kernel size. 3) The uniform baseline (\textbf{Uniform}): A potential drawback of the blurred baseline is its bias toward highlighting high-frequency information. Pixels that are very similar to their neighbors may receive less importance than pixels that differ significantly from their neighbors. To address this, missingness can be modeled by sampling a random uniform image (or activation) within the valid pixel range and using it as a baseline. 4) The normal baseline (\textbf{Normal}): This baseline is randomly drawn from a Normal distribution centered on the original image (or activation) with a specified variance parameter. 5) The training distribution baseline (\textbf{Train Data}): This method involves drawing instances from the training data and using them as baselines to generate multiple explanation maps. These maps are then averaged to produce a final explanation map (similar to Expected Gradients~\cite{erion2021improving}). In BEE, each baseline is formed by using the representation produced by the specific layer of interest. Detailed implementation specifics for each baseline type are provided in the Appendix.

In~\cite{sturmfels2020visualizing}, the authors explored the impact of various types of baselines including the aforementioned ones. Additionally, they investigated different baseline aggregation and averaging techniques. However, their findings did not indicate a specific baseline type as the optimal choice. Consequently, we posit that the richer and more comprehensive the baseline distribution $\mathcal{D}$, the higher the probability of sampling the a better-performing baseline. To achieve this goal, we propose to construct $\mathcal{D}$ as a mixture of distributions encompassing multiple types of baselines. In this manner, the process involves first drawing the baseline type according to the mixture weight and subsequently drawing the baseline from the distribution specific to that type.

While a straightforward way would be to set equal weights for all mixtures, we propose to learn a distribution for each mixture weight using the BEE approach that employs contextual EE of the baseline: At each iteration, the baseline type is sampled according to the current learned distribution of the mixture weights. This is followed by sampling a baseline of the specific distribution type, generating an explanation map, extracting a reward based on the explanation metric, and updating the distribution of mixture weights accordingly. This process enables the probabilistic selection of the most promising baseline type w.r.t. the specific context (the input $\x$) and explanation metric at hand $s$. In what follows, we describe this process in detail.

Let $c_{\theta}(\x) \in \R^K$ represent the context obtained by applying an auxiliary neural network $c_{\theta}$ (parameterized by $\theta$) to $\x$. The role of the function $c_{\theta}$ is to offer information about the particular input $\x$, thereby injecting context into the EE process. This adaptation allows the baseline distribution to adjust based on both the specific input $\x$ and the metric of interest.

Let $\mathcal{B}$ be a set of baseline types (e.g., Constant, Blur, etc.). Each baseline type $b \in \mathcal{B}$ is associated with a random vector $\w^b \in \R^K$ which follows a normal distribution with mean $\g^b$ and diagonal precision matrix represented by a vector $\q^b$. It is important to clarify that $\w^b$ does not represent the distribution of the baseline type $b$, but rather a random vector serving as a classifier (hyperplane) associated with the specific baseline type $b$. During the BEE procedure, each $\w^b$ is optimized based on the reward $r_b$ obtained from sampling a baseline of type $b$. The reward $r_b\in \{1,-1\}$ is modeled as a two-point distribution. Specifically $\text{Pr}(r_b|c_{\theta}(\x),\w^b)=\sigma(r_b c_{\theta}(\x) \cdot \w^b)$, where $\sigma(a)=(1+\exp(-a))^{-1}$ and $\cdot$ denotes the dot-product. Therefore, the dot-product between the context $c_{\theta}(\x)$ and the learned classifier $\w^b$ serves as the  (logit) score for the action of sampling a baseline of type $b$.

This formulation falls within the contextual EE framework: At each step, we play an action, i.e., sampling a baseline type $b$ among $\mathcal{B}$ (under a specific context $c_{\theta}(\x)$), drawing a baseline $\b$ from the selected distribution type\footnote{Note the distinction: $\b$ represents the baseline \textbf{representation}, while $b\in \mathcal{B}$ denotes the \textbf{type} of the baseline distribution (e.g., Blur, Uniform, etc.) from which $\b$ is sampled.} $b$, producing an explanation map $\m$, computing a reward (based on the specific metric of interest), and accordingly updating the relevant set of learnable parameters $\g^b, \q^b$ and $\theta$ s.t. the cumulative reward is maximized. It is important to emphasize that the BEE process is employed for each metric separately, enabling optimization per specific metric of interest.

At the beginning of the process, all $\g^b$ and $\q^b$ are set to $\mathbf{0}$ and $\mathbf{1}$, respectively (following the standard normal distribution). Then, at each iteration, given a context $c_{\theta}(\x)$, the following steps are performed:

\begin{enumerate}
    \item For each $z\in \mathcal{B}$, draw $\w^z$ from a normal distribution (using $\g^z$ and $\q^z$) and set
    $b \leftarrow \underset{z\in \mathcal{B}}{\text{argmax}} \,\,\sigma(c_{\theta}(\x) \cdot \w^z)$.

    \item Draw a baseline $\b$ from the baseline distribution of type $b$, compute an explanation map $\m$ according to Eq.~\ref{eq:abs-map} using $\b$ and $\x$.

    \item Compute the metric score $s(\m)$ and extract a corresponding reward $y$.
    
    \item $\u^*, \theta^* \leftarrow \underset{\u,\theta}{\text{argmin}} -\log \sigma(y \u \cdot c_{\theta}(\x)) +\frac{1}{2}\sum_{i=1}^K \q^b_i(\u_i-\g^b_i)^2$.

    \item $\g^b \leftarrow \u^*, \theta \leftarrow \theta^*$.
    
    \item $\q^b_i \leftarrow \q^b_i + \sigma(\g^b \cdot c_{\theta}(\x))\sigma(-\g^b \cdot c_{\theta}(\x))c_{\theta}(\x)^2_i$.
\end{enumerate}

Step 1 selects the baseline type $b$ with the highest expected reward. Step 2 draws a baseline $\b$ from the selected baseline distribution of type $b$, and uses it to form an explanation map $\m$. Step 3 produces a reward $y$ based on the metric score. For metrics that output binary explanation scores, the reward is simply mapped to 1 or -1 depending on success / failure. For metrics that outputs continuous scores, we compute the normalized rank $h\in [0,1]$ of the produced score, which is computed relative to the scores obtained from previous iterations. Then, the reward $y$ is drawn from a two-point distribution variable ($\{1,-1\}$) with a success parameter $h$. Steps 4-5 solve an optimization problem and update the mean $\g^b$ and the parameters of the context network $\theta$. The first term in the objective in Step 4 is a the likelihood of the reward given the model parameters and the context. The second term is a Gaussian prior serving as a proximal regularization on the update of the mean. The optimization in Step 4 is carried out by gradient descent w.r.t. $\u$ and $\theta$. Finally, the update of the precision $\q^b$ takes place in Step 6 and follows from the Laplace approximation. Once the learning process of $\mathcal{D}$ is completed, drawing a baseline is a straightforward process accomplished by applying Steps 1 and 2 in the above algorithm.

\subsection{BEE pretraining and finetuning}
\label{sec:abs-pretrain-finetune}
In practice, the BEE procedure outlined in Sec.~\ref{sec:abs} can be employed in two distinct phases: a mandatory pretraining phase followed by an \textbf{optional} finetuning phase. During the pretraining phase, we utilize a training set comprising instances from either the training or validation dataset. This phase involves training $c_{\theta}$ and pretraining $\g^b$ and $\q^b$ by iteratively applying the BEE procedure described in Sec.~\ref{sec:abs} for each instance in the training set. Specifically, in each epoch, we iterate over the instances in the training set and perform a single update to $\g^b$, $\q^b$, and $\theta$ based on the obtained reward. The pretraining phase is conducted offline and culminates in the optimization of the mixture of baseline distributions $\mathcal{D}$. Subsequently, when presented with a test instance, we employ the procedure described in Sec.~\ref{sec:explanation-map-construction} to obtain the most effective explanation map. This involves sampling $T$ baselines from $\mathcal{D}$, computing a pool of $T$ corresponding explanation maps using Eq.~\ref{eq:abs-map}, and selecting the best-performing explanation map based on the metric of interest, as expressed in Eq.~\ref{eq:abs-map-final}. For further enhancement, BEE finetuning can be employed during inference: Given the test instance $x$, online updates are applied to the pretrained baseline distribution $\mathcal{D}$ by reapplying the BEE procedure, refining $\g^b$ and $\q^b$ specifically for $x$ and the metric at hand (while keeping $\theta$ frozen). Therefore, the finetuning phase facilitates ongoing adaptation of the baseline distribution $\mathcal{D}$ to the characteristics of the test instance during the inference process. It is essential to highlight that the finetuning phase is optional. In the absence of finetuning, $\mathcal{D}$ retains the distribution optimized during the pretraining phase and remains static during inference; that is, $\q^b$ are not subject to further updates. The advantage of maintaining a static distribution lies in the ability to sample multiple baselines in parallel, as $\mathcal{D}$ remains unchanged across samples. Conversely, the finetuning phase necessitates sequential sampling, as $\mathcal{D}$ is refined after each sample based on the extracted reward. In Sec.~\ref{sec:experiments}, we comprehensively evaluate BEE in both settings (pretrained and finetuned). Our findings demonstrate that both versions yield state-of-the-art results, with the finetuning phase leading to additional improvements, albeit with an increase in runtime. The complexity of BEE with finetuning depends on the number of samples drawn from the learned baseline distribution. Sequential sampling is required for updates, adding to the overall complexity. In contrast, the complexity of BEE without finetuning is negligible due to the parallelization of the sampling process, and is comparable to the runtime of IG. A detailed theoretical analysis comparing the complexity of BEE to IG is provided in the Appendix.

\subsection{BEE implementation for CNN and VIT models}
\label{sec:impl}
In CNNs, $f$ comprises residual blocks~\cite{he2016resnet}, generating 3D tensors representing the activation maps $\x^l$. Additionally, $\psi$ (Eq.~\ref{eq:abs-map}) is set to the elementwise product, and $u$ averages across the channel axis to obtain a 2D map. The resulting map is then resized via bicubic interpolation to match the spatial dimensions of $\x$. In ViTs~\cite{dosovitskiy2020image}, $f$ consists of transformer encoder blocks, each associated with attention matrices. In this work, we opt to interpolate on the attention matrices. Therefore, we overload the notation and treat $\x^l$ as a 3D tensor in which each channel corresponds to an attention matrix in the $l$-th layer. Once the baseline matrices are drawn, they are further normalized by softmax to conform to probability distributions. $\psi$ is set to the Gradient Rollout (GR) - a variant of Attention Rollout~\cite{abnar2020quantifying}, in which the attention matrices are elementwise multiplied by their gradients. Detailed implementation of GR is provided in our Appendix and on our GitHub repository. The output of GR is the first row of a matrix corresponding to the \textsc{[CLS]} token. $u$ processes the output by truncating its initial element, reshaping it into a $14 \times 14$ matrix, and resizing to match the spatial dimensions of the input (with bicubic interpolation). Finally, the architecture of the context network $c$ was set to be a clone of the backbone of $f$, and was finetuned according the BEE procedure from Sec.~\ref{sec:abs}. The exact implementation of BEE for both architectures can be found in our GitHub repository.

%% file: latex/4_experiments.tex
\section{Experiments}
\label{sec:experiments}
Our experiments aim to address the following research questions (RQs): 
1) Does the BEE method outperform state-of-the-art methods? 2) Does BEE finetuning improve upon pretraining? 3) Do different metrics favor different explanation maps and baselines? 4) How does the number of sampled baselines $T$ affect BEE performance? 5) How does the performance of adaptive baseline sampling compare to non-adaptive sampling? 6) Does the learned baseline distribution obtained by BEE converge to the best-performing baseline distribution per metric? 7) Does integration on intermediate representation gradients improve upon integration on input gradients? 8) What is the contribution from context modeling in BEE? 9) Can other path-integration methods benefit from BEE? 
The primary manuscript addresses RQs 1-6 comprehensively. Specifically, RQs 1-2 are addressed in Tabs.~\ref{tab:cnn_backbones_metrics_exp} and \ref{tab:vit_backbones_metrics_exp}, RQ 3 is addressed in Tab.~\ref{tab:finetune_baselines_main} and Fig.~\ref{fig:graphs}, and RQs 4-6 are addressed in Fig.~\ref{fig:graphs}. Due to space limitations, experiments addressing RQs 7-9, along with additional analyses and ablation studies, are provided in the Appendix.

\subsection{Experimental setup}
\label{subsec:setup}
The experiments were conducted on an NVIDIA DGX 8xA100 Server. Our evaluation includes five model architectures:  ResNet101 (\textbf{RN})~\cite{he2016resnet}, DenseNet201 (\textbf{DN})~\cite{Huang2017DenselyCC}, ConvNext-Base (\textbf{CN})~\cite{Liu2022ACF}, ViT-Base (\textbf{ViT-B}) and ViT-Small (\textbf{ViT-S})~\cite{dosovitskiy2020image}.

\paragraph{Objective Evaluation Metrics} We conducted an extensive objective evaluation using a comprehensive set of explanation metrics to assess the faithfulness of the generated explanations. This faithfulness evaluation reveals the actual elements in the input the model relies on for its prediction. We consider the following set of metrics: the Area Under the Curve (AUC) of Positive (\textbf{POS}) and Negative (\textbf{NEG}) perturbations tests~\cite{chefer2021transformer}, AUC of the Insertion (\textbf{INS}) and Deletion (\textbf{DEL}) tests~\cite{petsiuk2018rise}, AUC of the Softmax Information Curve (\textbf{SIC}) and Accuracy Information Curve (\textbf{AIC})~\cite{kapishnikov2019xrai}, Average Drop Percentage (\textbf{ADP}) and Percentage Increase in Confidence (\textbf{PIC})~\cite{chattopadhay2018grad}. For POS, DEL, and ADP the lower the better, while for NEG, INS, SIC, AIC, and PIC the higher the better. It is important to clarify that while we report results for each metric according to its standard protocol, we also conducted experiments using various baselines for masking instead of the standard null baseline (a black image). Our findings indicate that the trends in the results remained consistent, regardless of the baseline used for masking. A detailed description of all metrics is provided in the Appendix.

\paragraph{Datasets} In accordance with previous works~\cite{kapishnikov2019xrai,kapishnikov2021guided,xu2020attribution,chefer2021transformer} we use the ImageNet~\cite{imagenet} ILSVRC 2012 (\textbf{IN}) validation set as our test set, which contains 50,000 images from 1,000 classes.

\paragraph{Methods} We consider a comprehensive set of explanation methods, covering gradient-based approaches, path-integration techniques, as well as gradient-free methods. Specifically, explanations for CNN models are generated by the following methods: Grad-CAM (\textbf{GC})~\cite{selvaraju2017grad}, Grad-CAM++ (\textbf{GC++})~\cite{chattopadhay2018grad}, Iterated Integrated Attributions (\textbf{IIA})~\cite{Barkan_2023_ICCV}, FullGrad (\textbf{FG})~\cite{Srinivas2019FullGradientRF}, Ablation-CAM (\textbf{AC})~\cite{Desai2020AblationCAMVE}, Layer-CAM (\textbf{LC})~\cite{jiang2021layercam}, LIFT-CAM (\textbf{LIFT})~\cite{Jung2021TowardsBE}, Integrated Gradients (\textbf{IG})~\cite{SundararajanTY17}, Guided IG (\textbf{GIG})~\cite{kapishnikov2021guided} and Blur IG (\textbf{BIG})~\cite{xu2020attribution}.
For ViT models, we consider the following state-of-the-art explanation methods: Transformer Attribution (\textbf{T-Attr})~\cite{chefer2021transformer}, Generic Attention Explainability (\textbf{GAE})~\cite{chefer2021generic} and IIA (which is applicable both for CNN and ViT architectures). Hyperparameters for all methods were configured according to the recommended settings published by the authors. A detailed description of all explanation methods is provided in the Appendix. Finally, our BEE method is evaluated in two modes: finetuned (\textbf{fBEE}) and pretrained (\textbf{pBEE}). For the pretraining phase, we used a separate training set of 5000 examples taken from the IN training set, avoiding overlap with the validation set used as a test set. Unless stated otherwise, we sampled $T=8$ baselines per test instance, and $n=10$ interpolation steps in the integration process (Eq.~\ref{eq:abs-map}). The integration was employed on the last convolutional / attention layer, i.e., we set $\mathcal{I} = \{L\}$ (Eq.~\ref{eq:abs-map-final}). A comparison of various settings of $\mathcal{I}$, including $L-1$ and $L-2$ is presented in the Appendix. The dimension of the context representation $K$ was set to match the output dimension of each backbone separately. Optimization in both the pretraining and finetuning phases was carried out using the Adam optimizer. For precise optimization details, please refer to the Appendix and our GitHub repository.

\subsection{Results}
\label{subsec:results}

\input{latex/tables/cnn_explanation_short}
\input{latex/tables/vit_explanation}

\begin{figure}[t]
\centering
    \includegraphics[width=0.8\columnwidth]{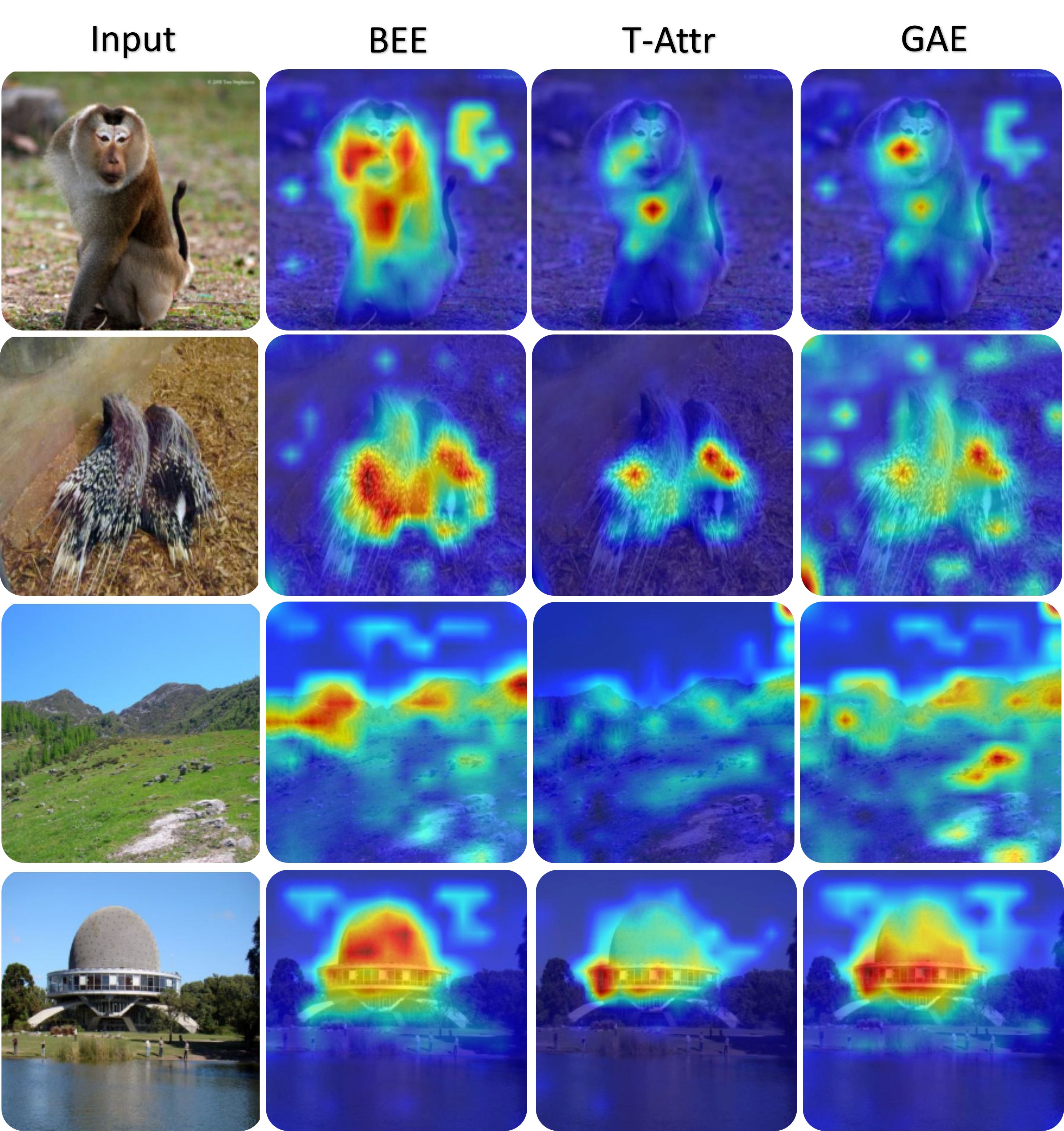}
    \caption{ViT qualitative results: Explanation maps produced using ViT-B w.r.t. the classes (top to bottom): ‘macaque’, ‘porcupine, hedgehog’, ‘alp’ and ‘planetarium’.}
    \label{fig:vitqr}
\end{figure}

Tables~\ref{tab:cnn_backbones_metrics_exp} and \ref{tab:vit_backbones_metrics_exp} present a quantitative comparison of fBEE, pBEE, and other state-of-the-art explanation methods on RN and ViT-B models, respectively. The full results, including CN and DN models as well as results for the ViT-S model, are presented in Tables~\ref{tab:cnn_backbones_metrics_exp_appendix} and \ref{tab:vit_backbones_metrics_exp_appendix} in Appendix~\ref{sec:explanation_tests_appendix}. The results indicate that fBEE is the top-performing method, with pBEE as the runner-up. The fact both fBEE and pBEE consistently outperform all other methods across all metrics underscores the effectiveness of BEE in explaining vision models. Figure~\ref{fig:vitqr} present qualitative comparisons of BEE against other explanation methods for the ViT-B model. Additionally, Figure~\ref{fig:cnnqr} in the Appendix provides qualitative results for the RN model. Arguably, BEE produces the most focused and class-discriminative explanation maps. These results correlate with the quantitative trends from Tables~\ref{tab:cnn_backbones_metrics_exp} and \ref{tab:vit_backbones_metrics_exp}.

\input{latex/tables/finetune_baselines}

Table~\ref{tab:finetune_baselines_main} presents the win-rate distribution (after pretraining) for each combination of baseline type and metric, utilizing the RN model. Each column displays the normalized \emph{win-rate} for each of the five baseline types relative to a specific metric. The win-rate was calculated by counting the instances where a particular baseline resulted in the best explanation map for the given metric, followed by normalization. Additionally, Fig.~\ref{fig:dists_appendix}, in the Appendix, present the reward distribution for each combination of baseline type and metric. The reward distributions were estimated using Monte Carlo approximation to the distribution of $\sigma(c_{\theta}(\x) \cdot \w^b)$ (by resampling from $\w^b$ for each $b\in \mathcal{B}$).
Notably, different metrics exhibit preferences for distinct types of baselines. For instance, PIC, INS, and NEG favor the Normal baseline, while DEL, POS, and SIC favor the Uniform baseline. These analyses demonstrate the need for adaptive adjustment of the baseline, as implemented by BEE.

\begin{figure}
\centering
    \includegraphics[width=0.48\textwidth]
    {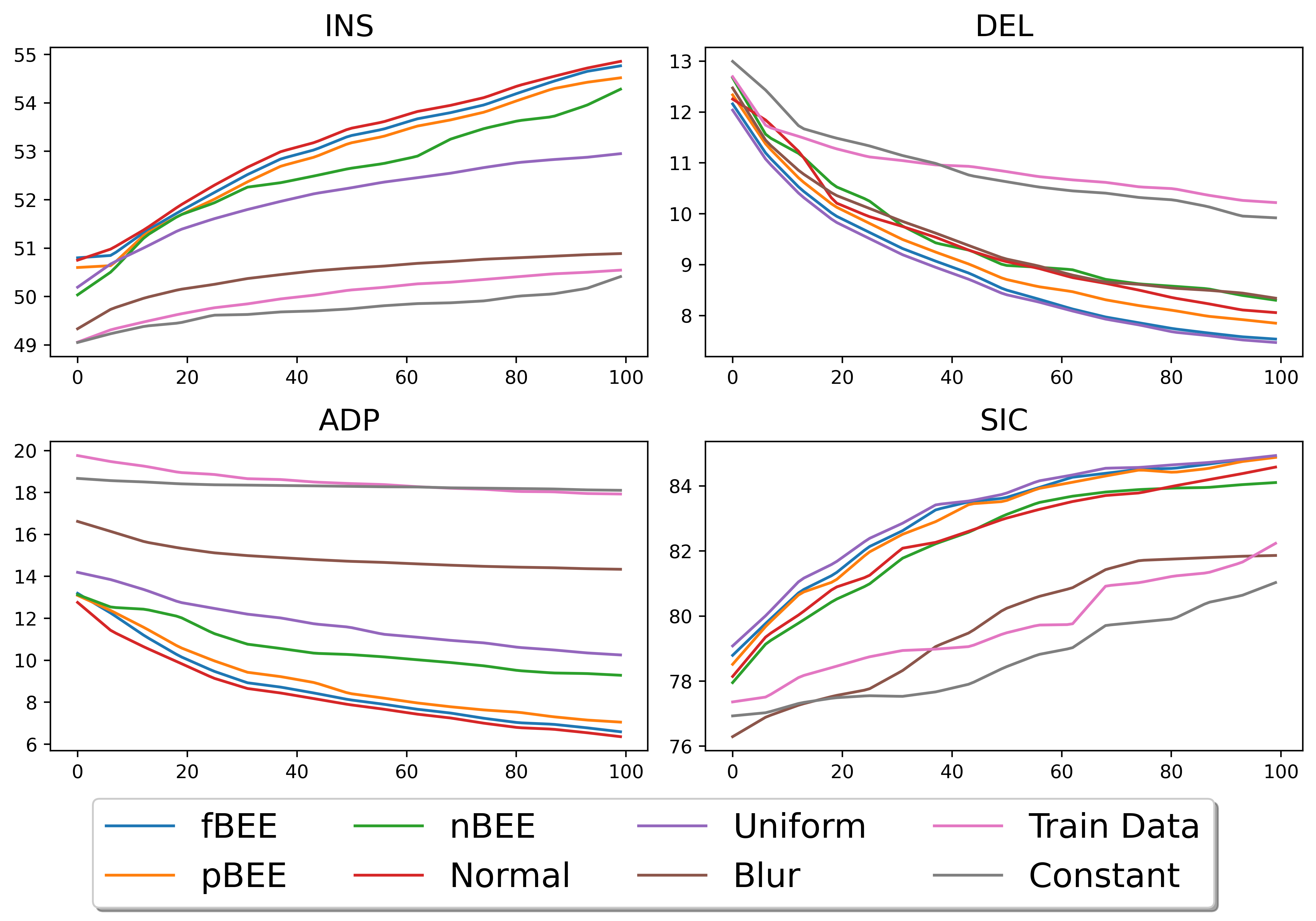}
    \caption{Metric score vs. number of drawn baselines. A comprehensive comparison among fBEE, pBEE, nBEE, and various types of baseline distributions is presented for each metric using the RN model.
    }
    \label{fig:graphs}
\end{figure}

Finally, we conducted an in-depth analysis of the performance of fBEE and pBEE in comparison to each distinct type of baseline distribution (Normal, Uniform, Train Data, Blur, and Constant), as the number of drawn baselines increases. Using a distinct type of baseline distribution is a special case in which the distribution on the types (mixture weights) conforms to the delta distribution on the specific type, i.e., consistently sampling from the same distinct type of baseline distribution. For completeness, we further consider the opposite extreme, by introducing a non-adaptive baseline sampling strategy (\textbf{nBEE}). This strategy involves uniformly drawing a baseline from each of the five distinct baseline distributions.
Following the analysis from Tab.~\ref{tab:finetune_baselines_main}, we observed that different metrics favor different distinct baselines. For example, the ADP metric favors the Normal baseline. Therefore, if we had an oracle that let us know a priori which is the best baseline type (per combination of input and metric), we could immediately choose it from the beginning. In the following analysis, we aim to investigate the rate of convergence of BEE to the best-performing baseline type, empirically. To this end, each sampling method was executed for 100 iterations, thereby generating 100 baselines. At each iteration, the best-performing baseline (among those sampled so far) was retained, resulting in weakly monotonic graphs. We replicated the experiment with 3000 test examples and reported the mean graph for each method.
Figure~\ref{fig:graphs} presents the results for BEE, pBEE, nBEE, and the rest of the distinct types of baseline distribution, using the RN model, focusing on the INS, DEL, ADP, and SIC metrics. A comprehensive figure containing all objective evaluation metrics is available in the Appendix. The horizontal axis in the figure represents the number of samples drawn from the baseline distribution, while the vertical axis corresponds to the metric score. Again, we observe that different metrics favor different baseline types.
Notably, fBEE exhibit the fastest convergence to the performance of the best-performing baseline type, thanks to its adaptive nature that promotes the most effective baseline distribution through online updates of the learned mixture of baseline distributions. Interestingly, even though pBEE does not employ finetuning, it marginally underperforms fBEE, indicating that the baseline distribution learned in the pretraining phase is already sufficiently effective on average. In contrast, nBEE, being a non-adaptive method, does not favor any specific baseline type, as it uniformly samples from each type of baseline. Hence, nBEE performs significantly worse, which is expected, as it does not promote the best-performing baseline types and exhausts many samples on less promising baseline types indefinitely.
In summary, both fBEE and pBEE demonstrate rapid convergence to the results of the best-performing type of baseline distribution. The findings in Fig.~\ref{fig:graphs} suggest that increased sample size correlates with improved performance. Yet, even with $T=8$, both fBEE and pBEE significantly outperforms all other methods, as evident in Tabs.~\ref{tab:cnn_backbones_metrics_exp} and \ref{tab:vit_backbones_metrics_exp}.

\paragraph{Ablation Study}
Additional ablation studies and analyses can be found in the Appendix. These ablation studies investigate diverse configurations of BEE by varying the number of sampled baselines ($T$) and the number of interpolation steps ($n$). Additionally, the analyses in the Appendix underscore the benefits of integrating gradients from intermediate network representations (RQ7), highlight the advantage of incorporating context modeling in BEE (RQ8), and explore the merit of applying the BEE method to other path-integration methods (RQ9), revealing  corresponding enhancements in performance. 

%% file: latex/tables/cnn_explanation_short.tex
\begin{table}[t]
  \centering
   \scalebox{0.8}{
    \begin{tabular}
    {c c c c c c c c c}
    \toprule
     & \multicolumn{1}{c}{NEG} & \multicolumn{1}{c}{POS} & \multicolumn{1}{c}{INS} & \multicolumn{1}{c}{DEL} & \multicolumn{1}{c}{ADP} & \multicolumn{1}{c}{PIC} & \multicolumn{1}{c}{SIC} & \multicolumn{1}{c}{AIC} \\
    \midrule
    GC & 56.41 & 17.82 & 48.14 & 13.97 & 17.87 & 36.69 & 76.91 & 74.36 \\
    GC++ & 55.20 & 18.01 & 47.56 & 14.17 & 16.91 & 36.53 & 76.44 & 71.97 \\
    LIFT & 55.39 & 17.53 & 45.39 & 15.32 & 18.03 & 35.95 & 76.73 & 72.76 \\
    AC & 54.98 & 19.38 & 47.05 & 14.23 & 16.18 & 35.52 & 73.36 & 70.35 \\
    IG & 45.66 & 17.24 & 39.87 & 13.49 & 37.52 & 19.94 & 54.67 & 51.92 \\
    GIG & 43.97 & 17.68 & 37.92 & 14.18 & 35.28 & 18.72 & 55.04 & 53.38 \\
    BIG & 42.25 & 17.44 & 36.04 & 13.95 & 40.85 & 24.53 & 56.98 & 53.36 \\
    FG & 54.81 & 18.06 & 42.68 & 14.64 & 21.06 & 31.59 & 75.35 & 71.49 \\
    LC & 53.52 & 17.92 & 46.11 & 14.31 & 24.34 & 35.43 & 73.93 & 65.77 \\
    IIA & 56.29 & 16.62 & 48.01 & 13.18 & 12.79 & 42.96 & 78.52 & 75.49 \\
    pBEE & \underline{59.10} & \underline{13.69} & \underline{51.15} & \underline{11.19} & \underline{11.35} & \underline{48.22} & \underline{81.23} & \underline{78.45} \\
    fBEE & \textbf{59.38} & \textbf{13.47} & \textbf{51.73} & \textbf{10.42} & \textbf{11.09} & \textbf{48.86} & \textbf{81.51} & \textbf{79.21} \\
    \bottomrule
      \end{tabular}}
\caption{Results on the IN dataset for the RN backbone: For POS, DEL and ADP, lower is better. For NEG, INS, PIC, SIC and AIC, higher is better. See Sec.~\ref{subsec:results} for details. }
  \label{tab:cnn_backbones_metrics_exp}
    \end{table}

%% file: latex/tables/vit_explanation.tex
\begin{table}
\begin{center}
\scalebox{0.8}{
\begin{tabular}{@{}ccccccccc@{}}
\toprule
& NEG & POS & INS & DEL & ADP & PIC & SIC & AIC \\
\midrule
T-Attr & 54.16 & 17.03 & 48.58 & 14.20 & 54.02 & 13.37 & 68.59 & 61.34 \\
GAE & 54.61 & 17.32 & 48.96 & 14.37 & 37.84 & 23.65 & 68.35 & 57.92\\
IIA & 56.01 & 15.19 & 49.31 & 12.89 & 33.93 & 26.18 & 68.92 & 62.38\\
pBEE & \underline{58.19} & \underline{12.51} & \underline{51.13} & \underline{10.94} & \underline{29.05} & \underline{31.01} & \underline{71.22} & \underline{65.32} \\
fBEE & \textbf{58.35} & \textbf{12.17} & \textbf{51.36} & \textbf{10.76} & \textbf{28.43} & \textbf{32.14} & \textbf{71.45} & \textbf{66.81} \\ \bottomrule
\end{tabular}}
\end{center}
\caption{Results on the IN dataset using ViT-B: For POS, DEL, and ADP, lower is better. For NEG, INS, PIC, SIC, and AIC, higher is better. See Sec.~\ref{subsec:results} for details.}
\label{tab:vit_backbones_metrics_exp}
\end{table}

%% file: latex/tables/finetune_baselines.tex
\begin{table}[ht!]
  \begin{center}
   
  \scalebox{0.8}{
    \begin{tabular}{@{}lc@{}lc@{}lc@{}lc@{}lc@{}lc@{}}%
    \toprule
      & & \multicolumn{1}{l}{NEG} & \multicolumn{1}{l}{POS} & \multicolumn{1}{l}{INS} & \multicolumn{1}{l}{DEL} & \multicolumn{1}{l}{ADP} & \multicolumn{1}{l}{PIC} & \multicolumn{1}{l}{SIC} & \multicolumn{1}{l}{AIC} \\
    \midrule
        & \multirow{1}{*}{Normal}  &  \multicolumn{1}{l}{\textbf{0.46}} & \multicolumn{1}{l}{0.27}  & \multicolumn{1}{l}{\textbf{0.41}} & \multicolumn{1}{l}{0.19} &  \multicolumn{1}{l}{\textbf{0.35}}& \multicolumn{1}{l}{\textbf{0.56}}  & \multicolumn{1}{l}{0.29} & \multicolumn{1}{l}{0.36} \\

        & \multirow{1}{*}{Uniform}  &  \multicolumn{1}{l}{0.32}& \multicolumn{1}{l}{\textbf{0.39}} & \multicolumn{1}{l}{0.26} & \multicolumn{1}{l}{\textbf{0.51}} &  \multicolumn{1}{l}{0.28}& \multicolumn{1}{l}{0.33}  & \multicolumn{1}{l}{\textbf{0.42}} & \multicolumn{1}{l}{\textbf{0.47}} \\

        & \multirow{1}{*}{Blur}  &  \multicolumn{1}{l}{0.06}& \multicolumn{1}{l}{0.16}  & \multicolumn{1}{l}{0.09} & \multicolumn{1}{l}{0.24} &  \multicolumn{1}{l}{0.03}& \multicolumn{1}{l}{0.08}  & \multicolumn{1}{l}{0.18} & \multicolumn{1}{l}{0.06} \\

        & \multirow{1}{*}{Constant}  &  \multicolumn{1}{l}{0.12}& \multicolumn{1}{l}{0.05}  & \multicolumn{1}{l}{0.13} & \multicolumn{1}{l}{0.04} &  \multicolumn{1}{l}{0.16}& \multicolumn{1}{l}{0.02}  & \multicolumn{1}{l}{0.04} & \multicolumn{1}{l}{0.08} \\

        & \multirow{1}{*}{Train Data  }  &  \multicolumn{1}{l}{0.04}& \multicolumn{1}{l}{0.13}  & \multicolumn{1}{l}{0.11} & \multicolumn{1}{l}{0.02} &  \multicolumn{1}{l}{0.18}& \multicolumn{1}{l}{0.01}  & \multicolumn{1}{l}{0.07} & \multicolumn{1}{l}{0.03} \\
        \midrule
  \end{tabular}}
  \end{center}
  \caption{The win-rate distribution (after pretraining) across different metrics and distinct types of baseline distributions.}
  \label{tab:finetune_baselines_main}
\end{table}

%% file: latex/5_conclusion.tex
\section{Conclusion}
\label{sec:conclusion}
This work recognized the diversity in explanation metrics and baseline representations, highlighting that different metrics exhibit preferences for distinct explanation maps based on the utilization of various baseline types. To address this double diversity, we introduced BEE, a path-integration method that introduces randomness to the integration process by sampling the baseline from a learned mixture of distributions. This mixture is learned through a contextual exploration-exploitation procedure, enhancing performance on the specific metric of interest. BEE can be applied in pretrained (pBEE) and finetuned (fBEE) modes, with the latter continually updating the baseline distribution during inference. Extensive evaluations across various model architectures demonstrate the superior performance of BEE compared to state-of-the-art explanation methods on a variety of objective evaluation metrics. In the Appendix, we further discuss limitations of BEE and avenues for future research.

%% file: latex/6_appendix.tex
\clearpage

\textbf{\Large{Supplementary Materials for Explainability with Baseline Exploration-Exploitation}}\\

\section{Appendix Overview}
\label{sec:appendix_overview}
The appendix provides detailed information and supplementary results complementing the main paper. Specifically, Sec.~\ref{sec:implementation_details} elaborates on the implementation details pertaining to our experiments. Section~\ref{sec:explanation_tests_appendix} provide a comprehensive overview of the explanation tests discussed in Sec.~\ref{sec:experiments}. Section~\ref{sec:bee_finetune_phase} discusses on the importance of adaptive baseline sampling and the preference for different baseline types by different inputs.
Section~\ref{sec:ablation} provides further ablation studies elucidating the impact of various configuration choices and demonstrating the advantages of applying the BEE mechanism to different path-integration methods.
Section~\ref{sec:complex} provides a theoretical analysis of the computational complexity of BEE.
Section~\ref{sec:metric_desc} provides comprehensive elucidation of the evaluation metrics employed in this work.
Section~\ref{sec:base_desc} describes the explanation methods included in our evaluation.
Section.~\ref{sec:gr} provides a detailed description of the Gradient Rollout technique.
Finally, in Sec.~\ref{sec:limitations}, we discuss the limitations of BEE and avenues for future research.

\subsection{Research Questions Recap}
\label{subsec:rq_recap}

Our experiments aim to address the following research questions (RQs): 
\begin{enumerate}
    \item Does the BEE method outperform state-of-the-art methods?
    \item Does BEE finetuning improve upon pretraining? 
    \item Do different metrics favor different explanation maps and baselines?
    \item How does the number of sampled baselines $T$ affect BEE performance?
    \item How does the performance of adaptive baseline sampling compare to non-adaptive sampling?
    \item Does the learned baseline distribution obtained by BEE converge to the best-performing baseline distribution per metric?
    \item Does integration on internal representation gradients improve upon integration on input gradients?
    \item What is the contribution from context modeling in BEE?

    \item Can other path-integration methods benefit from BEE?.
\end{enumerate}

The primary manuscript addresses RQs 1-6 comprehensively. Specifically, RQs 1-2 are addressed in Tabs.~\ref{tab:cnn_backbones_metrics_exp} and \ref{tab:vit_backbones_metrics_exp}, RQ 3 is addressed in Tab.~\ref{tab:finetune_baselines_main} and Fig.~\ref{fig:graphs}, 
and RQs 4-6 are addressed in Fig.~\ref{fig:graphs} (RQ 4 is further addressed in Tab.~\ref{tab:samples_ablation}). Additionally, RQs 7-8 are addressed in Table~\ref{tab:ablation_study_table}, and RQ 9 is addressed in Table~\ref{tab:ablation_study_pi_table}.

\section{Implementation Details}
\label{sec:implementation_details}
In the following section, we provide further implementation details regarding the BEE method.
\subsubsection{Baseline types implementation details} The baseline types presented on Sec.~\ref{sec:abs} are implemented as follows:
\begin{itemize}
    \item \textbf{Normal} - A baseline tensor is sampled from the normal distribution with the mean set to the activation $\v$, and the variance is set to $\frac{\sigma}{\max(\v) - \min(\v)}$. Here, $\sigma$ is sampled uniformly from [0.1, 0.5] and max and min produce vectors of the same dimensions as $\v$ occupied with the channel-wise maximum and minimum in $\v$, respectively.
    \item \textbf{Uniform} - A baseline tensor is sampled from the uniform distribution on the interval between the minimum and maximum for each channel in $\v$.
    \item \textbf{Blur} - The sigma parameter, which governs the kernel size and thus the strength of the blurring operation, is stochastically sampled for each trial within a range from 0 to 50.
    \item \textbf{Constant} - The baseline tensor is generated by randomly sampling values within the range defined by the minimum and maximum values of each activation channel / attention head.
    \item \textbf{Training data baseline} - In this method, the explanation map is formed by averaging over 100 explanation maps. Each map is created by randomly drawing an image from the training set, computing the activation / attention tensor for this image, and using this activation / attention tensor as a baseline for the integration process.
\end{itemize}

\subsubsection{Optimization details} For both the pretraining and finetuning phases, we used the AdamW optimizer with the default configuration suggested in PyTorch: a learning rate of 0.001, $\beta_1=0.9$, $\beta_2=0.999$, and weight decay of 0.01. Additionally, we utilized the ReduceLROnPlateau learning rate scheduler, with a termination criterion set to halt after a minimum of 10 epochs if no improvement in loss is observed.

\subsubsection{BEE Explanation map construction}
Following the description in Sec.~\ref{sec:explanation-map-construction}, in this work, we construct the explanation map according to Eqs.~\ref{eq:abs-map} and \ref{eq:abs-map-final} with the following parameter configuration: we used the representation produced by the last layer in the model ($L$), and thus set $\mathcal{I}={L}$. While we acknowledge that including representations from preceding layers, e.g., $L-1, L-2$, can potentially improve results further (see Sec.~\ref{sec:ablation}), even using the last layer alone demonstrates that both fBEE and pBEE significantly outperform all other methods. The number of interpolation steps is set to $n=10$. Our findings indicate that higher values of $n$ result in marginal improvements with higher computational costs, as depicted in Tab.~\ref{tab:steps_ablation}. Furthermore, the number of baselines sampled per example, during the finetuning phase, was set to $T=8$. As evident in Tabs.~\ref{tab:cnn_backbones_metrics_exp} and \ref{tab:vit_backbones_metrics_exp} (main paper), this small number was sufficient to obtain state-of-the-art results with both pBEE and fBEE.

Finally, for the exact implementation, the reader is referred to our GitHub repository.

\section{Full Explanation Tests Results}
\label{sec:explanation_tests_appendix}
Tables~\ref{tab:cnn_backbones_metrics_exp_appendix} and \ref{tab:vit_backbones_metrics_exp_appendix} present a quantitative comparison of fBEE, pBEE, and other state-of-the-art explanation methods on CN, DN and ViT-S models, respectively. These results, together with the results from Tabs.~\ref{tab:cnn_backbones_metrics_exp} and \ref{tab:vit_backbones_metrics_exp} indicate the effectiveness of BEE in explaining vision models.

\input{latex/tables/cnn_explanation_appendix}
\input{latex/tables/vit_explanation_appendix}

\begin{figure*}
\centering
    \includegraphics[width=1\textwidth]
    {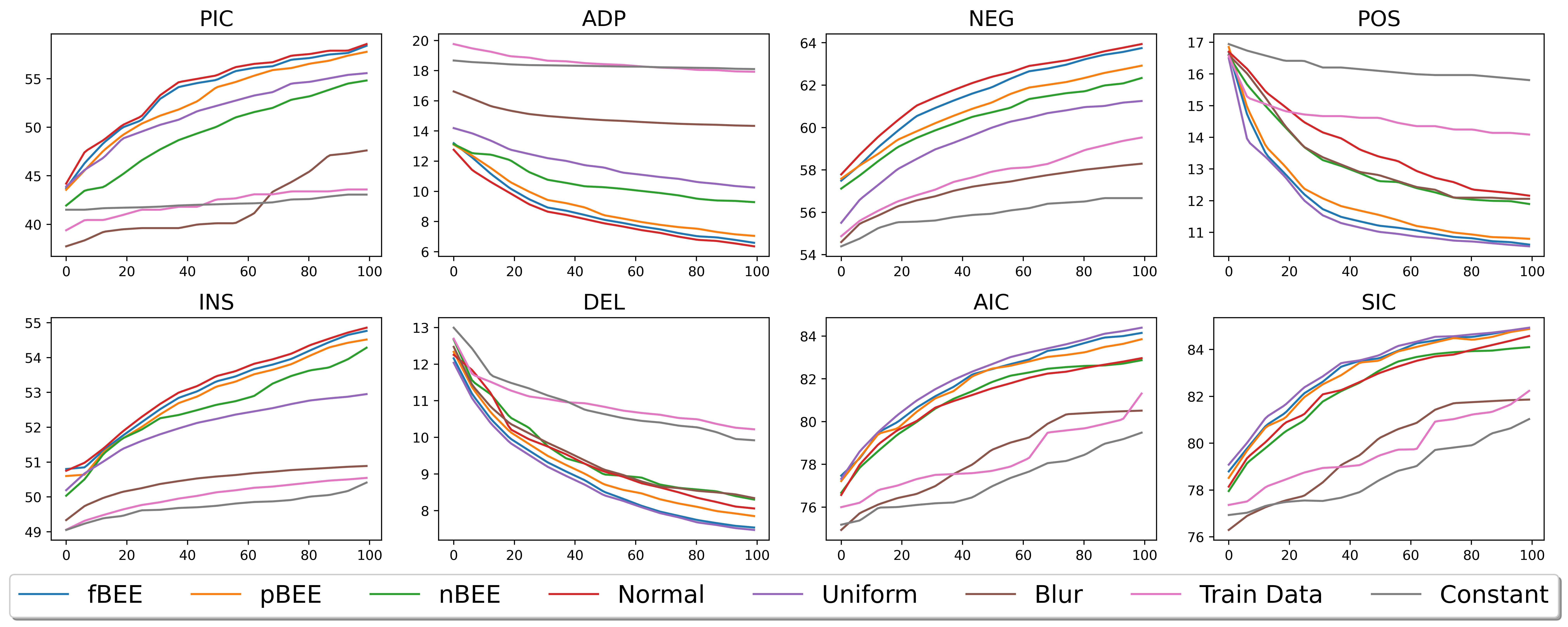}
    \caption{A comprehensive comparison among fBEE, pBEE, nBEE, and various types of baseline distributions is presented for each metric using the RN model. x axis - number of drawn baselines, y axis - metric score. See Section~\ref{subsec:results} for details.
    }
    \label{fig:graphs_appendix}
\end{figure*}

\begin{figure}
\centering
    \includegraphics[width=1.0\columnwidth]{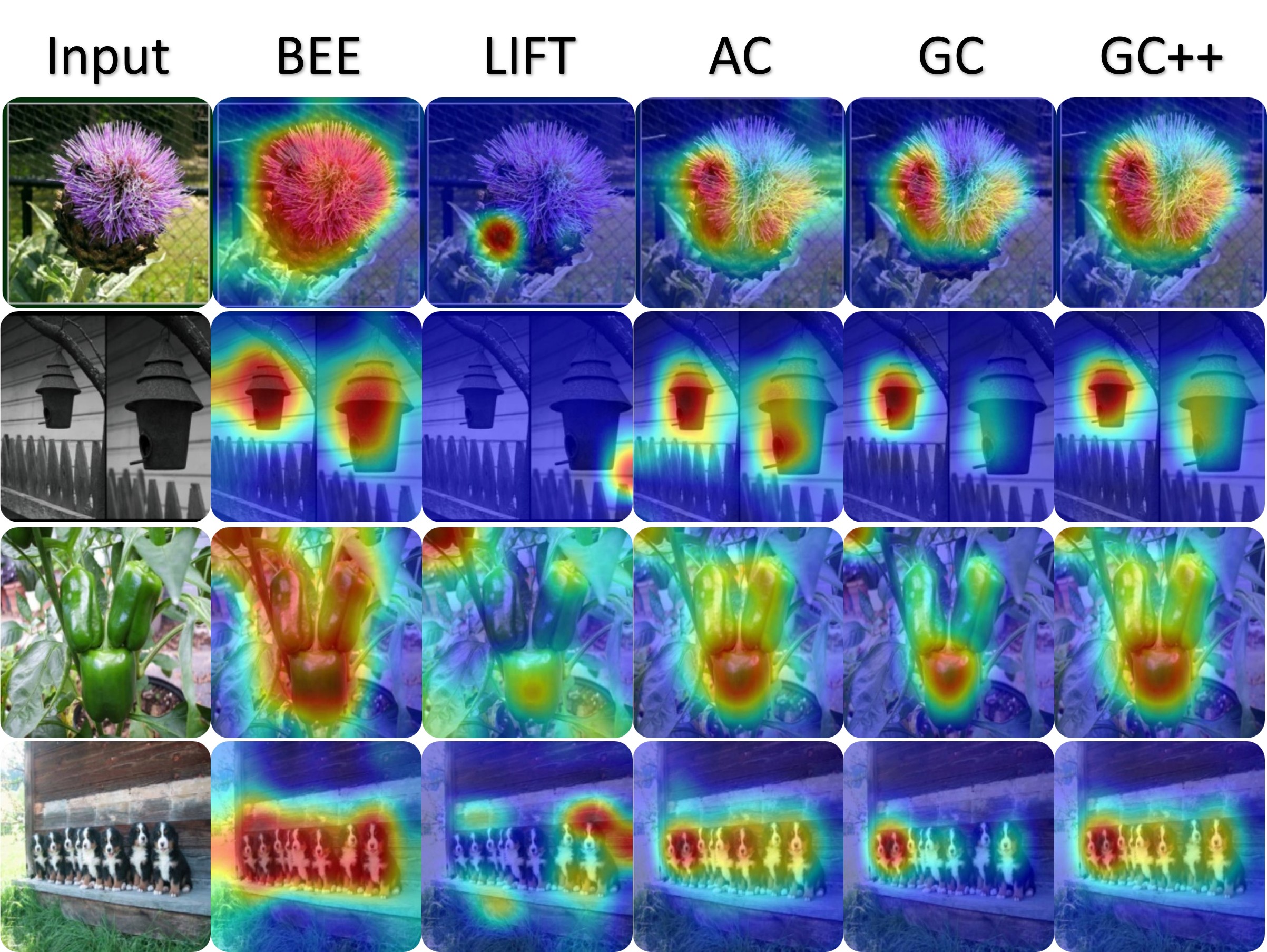}
    \caption{CNN qualitative results: Explanation maps produced
using RN w.r.t. the classes (top to bottom): ‘cardoon’, ‘birdhouse’, ‘bell pepper’, and ‘Appenzeller’.}
    \label{fig:cnnqr}
\end{figure}

Figure~\ref{fig:graphs_appendix} presents the complete comparison of baseline types referenced in Sec.~\ref{subsec:results}, encompassing all objective evaluation metrics. The figure illustrates consistent trends across the various metrics, further highlighting the observation that different metrics tend to favor different baseline types. Figure~\ref{fig:cnnqr} presents qualitative results for the RN model, further demonstrating BEE's ability to produce focused and accurate explanation maps.

\section{Importance of adaptive baseline sampling and the preference for different baseline types by different examples}
\label{sec:bee_finetune_phase}
 
\begin{figure}[t]
\centering
    \includegraphics[width=1.0\columnwidth]{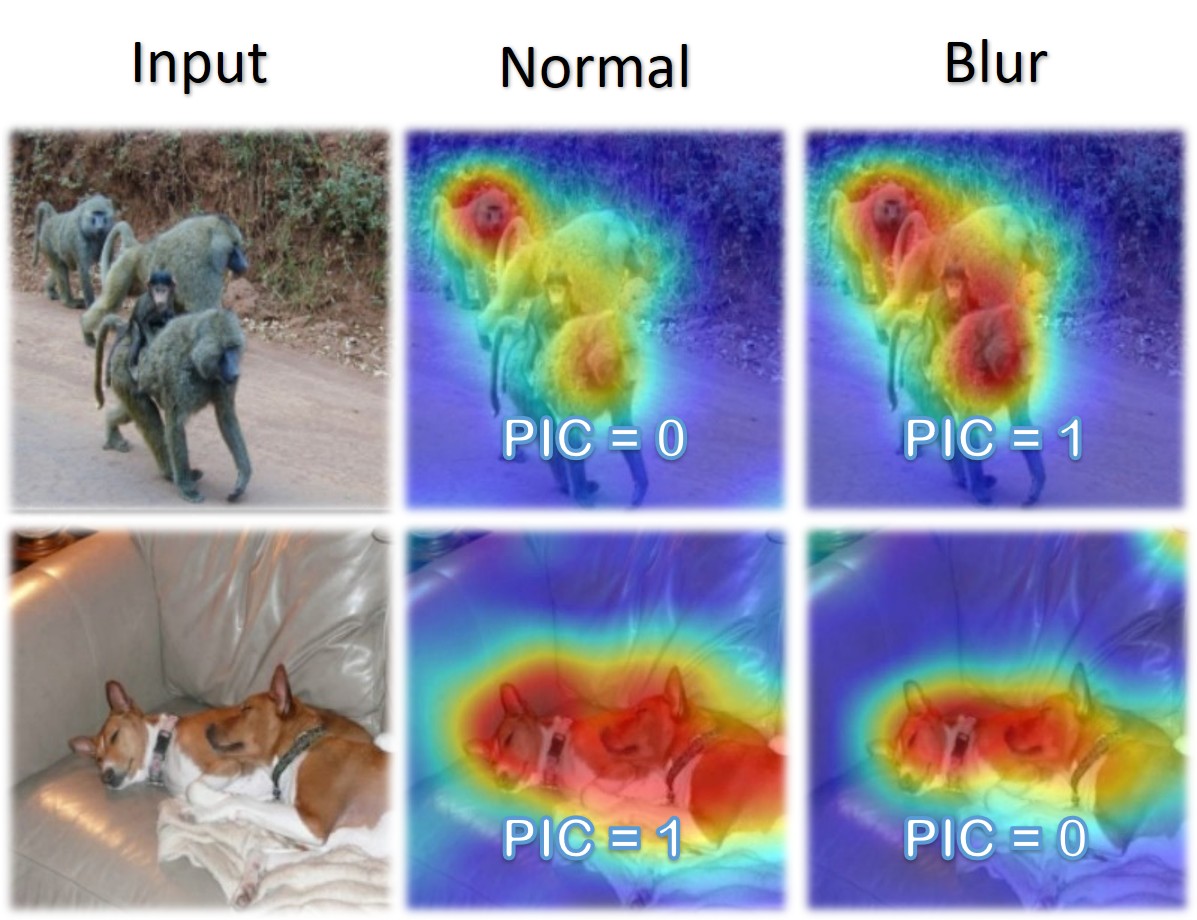}
    \caption{Explanation maps produced for the normal and blur baseline types using RN w.r.t. `baboon' and `basenji'. On the finetune phase over the PIC metric, BEE favored the normal baseline for the `baboon' and the blur baseline for the `basenji'. Both choices yielded better performance.}
    \label{fig:finetune_baselines_image}
\end{figure}

\begin{figure*}[ht!]
\centering
    \includegraphics[width=1\textwidth]{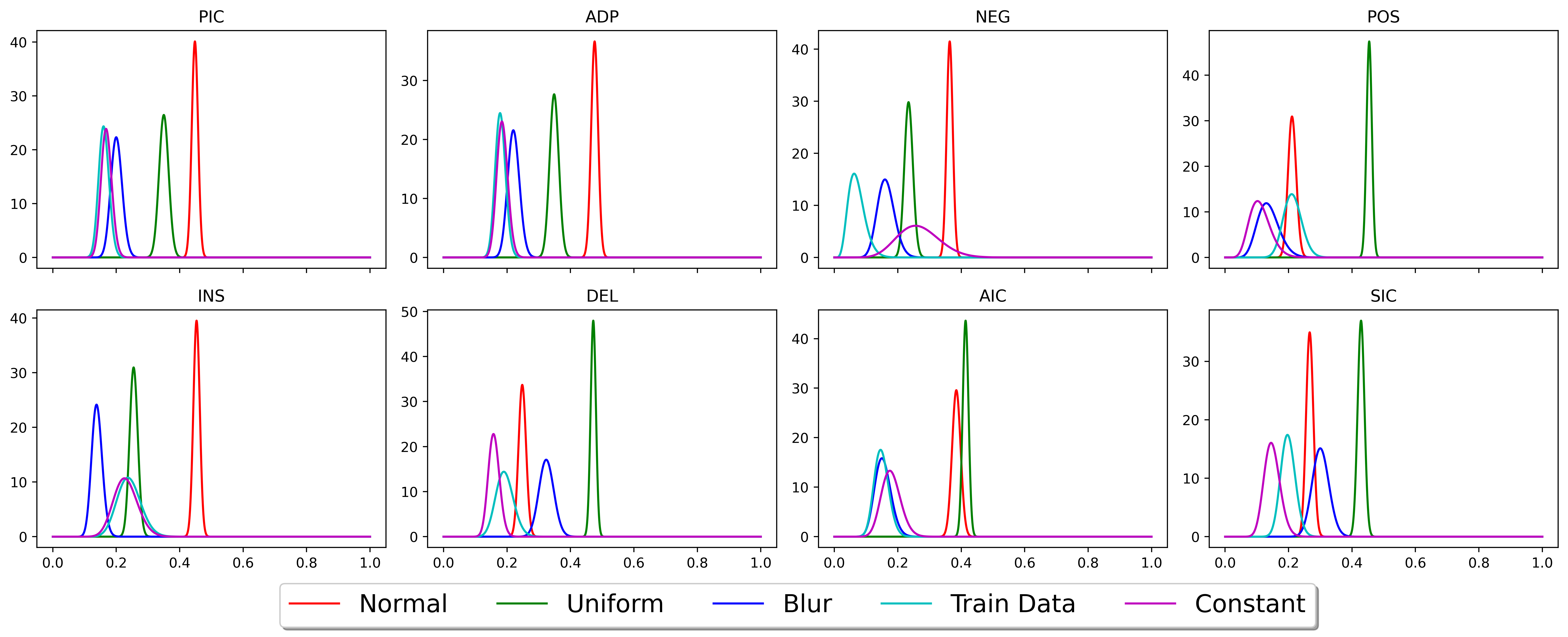}
    \caption{The reward distribution varies across different metrics and distinct types of baseline distributions. For example, PIC, INS, and NEG favor the Normal baseline, while DEL, POS, and SIC favor the Uniform baseline. The observed preferences of different metrics for distinct baselines underscore the necessity of the adaptive sampling employed by BEE.}
    \label{fig:dists_appendix}
\end{figure*}

Figure~\ref{fig:dists_appendix} presents the reward distribution (after pretraining) for each combination of baseline type and metric, utilizing the RN model. The reward distributions were estimated using Monte Carlo approximation to the distribution of $\sigma(c_{\theta}(\x) \cdot \w^b)$ (by resampling from $\w^b$ for each $b\in \mathcal{B}$). Notably, different metrics exhibit preferences for distinct types of baselines. For instance, PIC, INS, and NEG favor the Normal baseline, while DEL, POS, and SIC favor the Uniform baseline. This analysis provides another evidence for the effectiveness of BEE, showcasing its ability to promote the best-performing baseline type through adaptive baseline sampling.
Figure~\ref{fig:finetune_baselines_image} further underscores this statement, presenting explanation maps for two distinct samples of 'baboon' and 'basenji' classes. In the finetuning phase over the basenji sample, the optimal baseline selected for the PIC metric is of the Normal type. Conversely, for the baboon sample, the most effective baseline type is the Blur type. Both choices yielded better performance. These trends underscore the significance of the finetuning phase, wherein adaptations are made for each sample in separate. This is crucial, as distinct samples may exhibit preferences for different baseline types, even when evaluating the same metric.

\section{Ablation Study}
\label{sec:ablation}
In what follows, we present extensive ablation studies on several important design choices and parameters: (I) The contribution from integrating on the internal network representations and their gradients (as opposed to the input image gradients as done in IG). (II) The number of sampled baselines per test example (during the finetuning phase) - $T$. (III) The number of interpolation steps - denoted as $n$ in Eq.~\ref{eq:abs-map}. (IV) The contribution from integrating multiple layers.

We further underscore the advantages of employing BEE's adaptive baseline sampling with different path-integration methods such as Integrated Gradients (IG), Blur Integrated Gradients (BIG), and Guided Integrated Gradients (GIG). This emphasis aims to highlight that BEE's distinctive sampling approach is not exclusive to a specific path-integration method but extends its benefits to other counterparts as well.

\subsubsection{The contribution from integrating on the internal network representations and their gradients} Table~\ref{tab:ablation_study_table} presents comparison of three alternatives:
(1) IG-fBEE - This approach involves computing gradients with respect to the input image instead of the internal network representations. In this case, the baselines match the dimensions of the image.
(2) ACT-IG - A variant of Integrated Gradients (IG), in which the internal representations of the network are employed for the integration process instead of an input image. Specifically, the last layer's representation is utilized to implement this variation of IG. (3) non-contextual BEE (ncBEE) - A variant of BEE that does not incorporate information regarding the input (hence does not utilize the context network $c_{\theta}$). Instead, the reward for each baseline type $\r_b$ is modeled by a Beta distribution, and the alpha and beta parameters are estimated by counting success and failures (per metric).
(4) BEE (pBEE, fBEE) - The methods presented in the paper.
(1), (3), and (4) are configured with $n=10$ and $T=8$.
The results in Tab.~\ref{tab:ablation_study_table} substantiate the rationale for employing internal network representations and the benefit from using the contextual sampling approach. Notably, IG-fBEE demonstrates suboptimal performance across all metrics, except for POS and DEL, where it marginally trails BEE. Furthermore, we observe IG-fBEE achieving inferior results in comparison to ACT-IG in the majority of the metrics. This underscores the efficacy of leveraging internal representations and their gradients. Although ncBEE outperforms IG-fBEE and ACT-IG, it consistently demonstrates inferior performance compared to BEE, thereby underscoring the advantages of employing the contextual sampling approach.

\input{latex/tables/ablation_study}

\subsubsection{The contribution from applying BEE mechanism to other path-integration methods} Table~\ref{tab:ablation_study_pi_table} presents comparison of the following alternatives:
(1) IG-fBEE - This approach involves computing gradients with respect to the input image instead of the internal network representations. In this case, the baselines match the dimensions of the image.
(2) BIG-fBEE - This approach involves the same map construction process as in BIG, but with a baseline sampled in the same manner as in fBEE.
(3) GIG-fBEE - This approach involves the same map construction process as in GIG, but with a baseline sampled in the same manner as in fBEE.
(4) BEE (pBEE, fBEE) - The methods presented in the paper.
(1)-(4) are configured with $n=10$ and $T=8$.
The results presented in Tab.~\ref{tab:ablation_study_pi_table} reveal notable advantages resulting from the integration of BEE across various path-integration methods.
\input{latex/tables/ablation_study_bee_pi_smooth}

\subsubsection{The number of sampled baselines $T$}
Table~\ref{tab:samples_ablation} presents the results of fBEE and pBEE across varying values $T$ (sampled baselines). Notably, we observe that beginning from $T=8$, both fBEE and pBEE exhibit commendable performance.  Although higher values of $T$ hold the potential for improved results, our findings suggest that $T=8$ is adequate for achieving state-of-the-art performance while maintaining acceptable runtime.
It is worth mentioning that in Sec.~\ref{sec:complexity}, we discuss the idea of parallelizing pBEE computation on the GPU. Our examination reveals that, with sufficient computing resources, utilizing pBEE with $T>8$ becomes feasible, all while maintaining runtimes comparable to those of IG.
\input{latex/tables/ablation_samples}

\subsubsection{The number of interpolation steps $n$} Table~\ref{tab:steps_ablation} provides a quantitative and runtime analysis comparing various configurations of the interpolation step number $n$ across 1000 random samples from the IN dataset over RN.  We chose to evaluate over the ACT-IG method instead of BEE in order to prevent inconsistency arising from the stochastic nature of BEE. The comparison shows that the gain from using $n > 10$, if exists, is insignificant, i.e., setting $n$ = 10 is optimal.
\input{latex/tables/ablation_of_steps}

\section{Computational Complexity and Runtime Comparison}
\label{sec:complex}
In this section, we provide a theoretical analysis of the computational complexity of our BEE method in both modes (pretrained and finetuned). 
\label{sec:complexity}

For Integrated Gradients (IG), a singular forward-backward pass is executed during the explanation map creation process. Therefore, given that $n$ represents the number of interpolation steps and $B$ denotes the maximal batch size accommodated by the GPU, the computational complexity of IG is $\lceil\frac{n}{B}\rceil$. Assuming a GPU with $nT \leq B$, it holds that $\lceil\frac{nT}{B}\rceil \leq 1$, and hence the cost of pBEE is bounded by a \emph{single} forward-backward pass. As for fBEE, its complexity, relative to IG, is $\frac{\lceil\frac{n}{B}\rceil T}{\lceil\frac{n}{B}\rceil} = T$ due to the serial nature of the computations across different trials.
Practically, for pBEE with $n=10$ and $T=8$, it is sufficient to have $B=80$, to obtain $\lceil\frac{nT}{B}\rceil=\mathcal{O}(1)$, which is feasible with a 8xA100 GPU server (640GB RAM). In these cases, the runtimes of pBEE and IG are equivalent. In fact, if  $nT \leq B$, pBEE can become even faster than IG, since in IG the gradients are backpropagated through all layers back to the \emph{input}, while in this work pBEE gradients are backpropagated to the last layer $L$ only. Finally, one can easily distribute pBEE computation across several machines to obtain further speed-up.

\section{Evaluation Metrics}
\label{sec:metric_desc}

There is no single measure or test set which is generally acceptable for evaluating explanation maps. Hence, in order to ensure comparability, the evaluations in this research  follow earlier works~\cite{chattopadhay2018grad,chefer2021transformer,kapishnikov2019xrai,petsiuk2018rise}. In general, the various tests entail different types of masking of the original input according to the explanation maps and investigating the change in the model's prediction for the masked input compared to its original prediction based on the unmasked input. There are two variants for these tests which differ based on the class of reference. In one variant, the difference in predictions refers to the ground-truth class, and in the second variant, the difference in predictions refers to the model's original top-predicted class. In the manuscript, we report results for both variants and dub the first variant as `target' and the second variant as `predicted', respectively. 

In what follows, we list and define the different evaluation measures used in this research:
\begin{enumerate}
   \item {Average Drop Percentage \textbf{(ADP)}~\cite{chattopadhay2018grad}: $\text{ADP}=100\%\cdot\frac{1}{N} \sum_{i=1}^N \frac{\max(0,Y_i^c-O_i^c)}{Y_i^c},$ where $N$ is the total number of images in the evaluated dataset, $Y_i^c$ is the model's output score (confidence) for class $c$ w.r.t. the original image $i$. $O_i^c$ is the same model’s score, this time w.r.t. to a masked version of the original image (produced by the Hadamard product of the original image with the explanation map). The \textbf{lower} the ADP the better the result.}
   \item {Percentage of Increase in Confidence \textbf{(PIC)}~\cite{chattopadhay2018grad}: $\text{PIC} = 100\% \cdot \frac{1}{N} \sum_{i=1}^N {1} (Y_i^c < O_i^c)$. PIC reports the percentage of cases in which the model’s output scores increase as a result of the replacement of the original image with the masked version based on the explanation map. The explanation map is expected to mask the background and help the model to focus on the original image. Hence, the \textbf{higher} the PIC the better the result.}
   \item Perturbation tests entail a stepwise process in which pixels in the original image are gradually masked out according to their relevance score obtained from the explanation map~\cite{chefer2021transformer}. At each step, an additional 10\% of the pixels are removed and the original image is gradually blacked out. The performance of the explanation model is assessed by measuring the area under the curve (AUC) with respect to the model's prediction on the masked image compared to its prediction with respect to the original (unmasked) image. 
   We consider two types of masking:
   \begin{enumerate}
       \item {Positive perturbation (\textbf{POS}), in which we mask the pixels in decreasing order, from the highest relevance to the lowest, and expect to see a steep decrease in performance, indicating that the masked pixels are important to the classification score. Hence, for the POS perturbation test, lower values indicate better performance.}
       \item {Negative perturbation (\textbf{NEG}), in which we mask the pixels in increasing order, from lowest to highest. A good explanation would maintain the accuracy of the model while removing pixels that are not related to the class of interest. Hence, for the NEG perturbation test, lower values indicate better performance.}
   \end{enumerate}
   {In both positive and negative perturbations, we measure the area-under-the-curve (AUC), for erasing between \text{10\%-90\%} of the pixels.
As explained above, results are reported with respect to the `predicted' or the `target' (ground-truth) class.}
   \item{The deletion and insertion metrics ~\cite{petsiuk2018rise} are described as follows:}
   \begin{enumerate}
       \item {The deletion (\textbf{DEL}) metric measures a decrease in the probability of the class of interest as more and more important pixels are removed, where the importance of each pixel is obtained from the generated explanation map. A sharp drop and thus a low area under the probability curve (as a function of the fraction of removed pixels) means a good explanation.}
       \item {In contrast, the insertion (\textbf{INS}) metric measures the increase in probability as more and more pixels are revealed, with higher AUC indicative of a better explanation.}
   \end{enumerate}
   {In this work, we remove pixels by setting their value to zero. Gradual removal or introduction of pixels is performed in steps of 0.1 i.e., remove or introduce 10\% of the pixels on each step).}

   \item {The Accuracy Information Curve (\textbf{AIC}) and the Softmax Information Curve (\textbf{SIC})~\cite{kapishnikov2019xrai} metrics are both similar in spirit to the receiver operating characteristics (ROC). These measures are inspired by the Bokeh effect in photography~\cite{liu2016stereo}, which consists of focusing on objects of interest while keeping the rest of the image blurred. In a similar fashion, we start with a completely blurred image and gradually sharpen the image areas that are deemed important by a given explanation method. Gradually sharpening the image areas increases the information content of the image. We then compare the explanation methods by measuring the approximate image entropy (e.g., compressed image size) and the model’s performance (e.g., model accuracy).

}
    \begin{enumerate}
       \item { 
       
The AIC metric measures the accuracy of a model as a function of the amount of information provided to the explanation method. AIC is defined as the AUC of the accuracy vs. information plot. The information provided to the method is quantified by the fraction of input features that are considered during the explanation process.

}
       \item { 
       
The SIC metric measures the information content of the output of a softmax classifier as a function of the amount of information provided to the explanation method. SIC is defined as the AUC of the entropy vs. information plot. The entropy of the softmax output is a measure of the uncertainty or randomness of the classifier's predictions. The information provided to the method is quantified by the fraction of input features that are considered during the explanation process.

}
   \end{enumerate}
 \end{enumerate}

\section{Explanation Methods}
\label{sec:base_desc}
\begin{enumerate}
       \item {Grad-CAM (\textbf{GC})~\cite{selvaraju2017grad} integrates the activation maps from the last convolutional layer in the CNN by employing global average pooling on the gradients and utilizing them as weights for the feature map channels. 
       }
       \item {Grad-CAM++ (\textbf{GC++})~\cite{chattopadhay2018grad} is an advanced variant of Grad-CAM that utilizes a weighted average of the pixel-wise gradients to generate the activation map weights.
       }
       \item {Iterated Integrated Attributions (\textbf{IIA})~\cite{Barkan_2023_ICCV} an explanation approach that generalizes Integrated Gradients to an iterated integral.
       }
       \item {Integrated Gradients (\textbf{IG})~\cite{SundararajanTY17} integrates over the interpolated image gradients. }
       \item {Blur IG (\textbf{BIG})~\cite{xu2020attribution} is concerned with the introduction of information using a baseline and opts to use a path that progressively removes Gaussian blur from the attributed image. }
       \item {Guided IG (\textbf{GIG})~\cite{kapishnikov2021guided} improves upon Integrated Gradients by introducing the idea of an adaptive path method. By calculating integration along a different path than Integrated Gradients, high gradient areas are avoided which often leads to an overall reduction in irrelevant attributions.}
       \item {LIFT-CAM (\textbf{LIFT})~\cite{Jung2021TowardsBE} employs the DeepLIFT~\cite{Shrikumar2017LearningIF} technique to estimate the activation maps SHAP values~\cite{Lundberg2017AUA} and then combine them with the activation maps to produce the explanation map.}
       \item {The FullGrad (\textbf{FG}) method~\cite{Srinivas2019FullGradientRF} provides a complete modeling approach of the gradient by also taking the gradient with respect to the bias term, and not just with respect to the input.}
       \item {LayerCAM (\textbf{LC})~\cite{jiang2021layercam} utilizes both gradients and activations, but instead of using the Grad-CAM approach and applying pooling on the gradients, it treats the gradients as weights for the activations by assigning each location in the activations with an appropriate gradient location. The explanation map is computed with a location-wise product of the positive gradients (after ReLU) with the activations, and the map is then summed w.r.t. the activation channel, with a ReLU applied to the result.
       }
       \item {Ablation-CAM (\textbf{AC})~\cite{Desai2020AblationCAMVE} is an approach that only uses the channels of the activations. It takes each activation channel, masks it from the final map by zeroing out all locations of this channel in the explanation map produced by all the channels, computes the score on the masked explanation map (the map without the specific channel), and this score is used to assign an importance weight for every channel. At last, a weighted sum of the channels produces the final explanation map.
       }
       \item {The Transformer attribution (\textbf{T-ATTR})~\cite{chefer2021transformer} method computes the importance of each input token by analyzing the attention weights assigned to it during self-attention. Specifically, it computes the relevance score of each token as the sum of its attention weights across all layers of the Transformer. The intuition behind this approach is that tokens that receive more attention across different layers are likely more important for the final prediction.
To obtain a more interpretable and localized visualization of the importance scores, the authors also propose a variant of the method called Layer-wise Relevance Propagation (LRP), which recursively distributes the relevance scores back to the input tokens based on their contribution to the intermediate representations.}
       \item {Generic Attention Explainability (\textbf{GAE})~\cite{chefer2021generic} is a generalization of T-Attr for explaining Bi-Modal transformers.}
\end{enumerate}

\section{Gradient Rollout Implementation}
\label{sec:gr}
The Gradient Rollout (\textbf{GR}) technique is a modified version of the Attention Rollout (\textbf{AR})~\cite{abnar2020quantifying} method, which differentiates itself by including a Hadamard product between each attention map and its gradients in the computation, rather than relying solely on the attention map.
The GR method can be expressed mathematically as follows:
\begin{equation}
\label{eq:gr1}
A'_b=I+E_{h}(A_b \circ G_b),
\end{equation}
\begin{equation}
\label{eq:gr2}
GR=A'_1 \cdot A'_2 \cdot \cdot \cdot A'_B.
\end{equation}
where $A_b$ is a 3D tensor consisting of the 2D attention maps produced by each attention head in the transformer block $b$, $G_b$ is the gradients w.r.t. $A_b$.
$I$ is the identity matrix, $B$ is the number of transformer blocks in the model, $E_h$ is the mean reduction operation (taken across the attention heads dimension), and $\circ$ and $\cdot$ are the Hadamard product and matrix multiplication operators, respectively. Following this, GR proceeds with the original Rollout computation~\cite{abnar2020quantifying}, resulting in the first row of the derived matrix (associated with the [CLS] token). Finally, this output is processed by truncating its initial element and reshaping it into a 14 × 14 matrix. The exact implementation of GR appears in our GitHub repository.

\section{Limitations and Future Work}
\label{sec:limitations}

While the BEE method exhibits promising results showing enhanced explainability across all metrics, certain limitations and avenues for future exploration are acknowledged.

\subsubsection{Faster convergence to the optimal baseline}
The finetuning process of the BEE method (fBEE) introduces a sequential sampling approach, increasing computational complexity compared to the parallelized sampling in the pretrained BEE version (pBEE). Although pBEE demonstrates state-of-the-art results, fBEE exhibits an additional performance boost across all metrics.

The choice between pBEE and fBEE involves a discernible trade-off. For scenarios prioritizing speed without significant compromise on explanation precision, pBEE may be the preferred choice. Conversely, when precision is paramount and an increase in runtime is acceptable, fBEE emerges as the optimal solution.

As part of our future research we plan to explore techniques for accelerating the fBEE process and improving the pBEE prediction. Given that fBEE involves sequential sampling, investigating mechanisms that consider all baselines sampled so far within each type of baseline distribution is pertinent. Currently, when a baseline type is sampled, fBEE simply resamples from the distribution of that baseline type without incorporating knowledge of the baselines already drawn from this specific distribution throughout the finetuning process. Integrating such information as additional context could empower fBEE to focus on or avoid particular areas in the baseline space of the specific type, potentially expediting the finetuning steps by converging more rapidly to improved baseline representations within each type.

In essence, investigating methods to integrate contextual information about baselines sampled within a distribution during both the pretraining and finetuning processes shows promise for enhancing predictions by both pBEE and fBEE. This aligns with the broader objective of advancing the computational efficiency of sequential sampling methods, thereby optimizing the finetuning stage in the BEE framework. Nevertheless, providing better contextual information to pBEE, making it aware of less promising regions in the baseline space for the specific input under consideration, has the potential to significantly contribute to improved pBEE prediction as well.

\subsubsection{Optimization across multiple metrics simultaneously}
The evaluation in this work encompasses a diverse array of explainability metrics. These metrics represent established methodologies to quantify the performance of explanation methods, as commonly adopted by recent XAI research. Our evaluation framework adheres to these established protocols and metrics, demonstrating the BEE method's notable outperformance over other state-of-the-art methods across various model architectures.

The motivation behind the BEE methodology stems from the realization that each metric captures specific facets of model explainability, potentially promoting different explanations. However, practitioners often seek an optimal single explanation that excels across multiple metrics, while our BEE method enables the generation of explanations that perform optimally on individual metrics.

A logical progression from this juncture involves developing mechanisms that empower BEE to produce explanations excelling across a set of metrics rather than focusing on a single metric. This can be achieved through the design of complex reward functions capable of incorporating and balancing multiple metric scores simultaneously. Such advancements hold the potential to provide practitioners with holistic explanations that cater to diverse aspects of model explainability. Yet, it is crucial to note that careful consideration is needed in designing the reward function, as different metrics may promote conflicting goals. Nevertheless, until a consensus is reached on the ultimate explanation metric, we believe our proposed BEE method serves as a valuable tool for adapting explanations per metric, contributing to the ongoing discourse in the quest for robust and comprehensive model explainability.

\subsubsection{Generalization to Different Domains:} Our current evaluation primarily focuses on vision models. Extending the applicability of BEE to different domains, such as natural language processing and audio processing, presents an exciting avenue for future research.
\\

\noindent In conclusion, addressing the aforementioned limitations and exploring the suggested future avenues might contribute to the continuous development and refinement of the BEE method, making it a more robust and versatile machinery for model explainability across diverse applications.

%% file: latex/tables/cnn_explanation_appendix.tex
\begin{table*}[t]

\begin{center}
   \scalebox{0.9}{
    \begin{tabular}
    {l  l  | lc@{}lc@{}lc@{}lc@{}lc@{}lc@{}lc@{}}
    \toprule
      &  & \multicolumn{1}{c}{GC} & \multicolumn{1}{c}{GC++} & \multicolumn{1}{c}{LIFT} & \multicolumn{1}{c}{AC} & \multicolumn{1}{c}{IG}& \multicolumn{1}{c}{GIG}& \multicolumn{1}{c}{BIG} & \multicolumn{1}{c}{FG} & \multicolumn{1}{c}{LC} & \multicolumn{1}{c}{IIA} & \multicolumn{1}{c}{pBEE}& \multicolumn{1}{c}{fBEE}\\
    \midrule
    \multirow{8}{*}{CN}

    & \multirow {1}{*}{NEG}   & \multicolumn{1}{c}{52.86} & \multicolumn{1}{c}{53.82} & \multicolumn{1}{c}{53.98} & \multicolumn{1}{c}{53.68} & \multicolumn{1}{c}{45.24} &\multicolumn{1}{c}{41.43} & \multicolumn{1}{c}{40.72}& \multicolumn{1}{c}{52.06} & \multicolumn{1}{c}{54.12} & \multicolumn{1}{c}{55.94} & \multicolumn{1}{c}{\underline{58.19}}& \multicolumn{1}{c}{\textbf{58.77}}\\

    & \multirow {1}{*}{POS}   &  \multicolumn{1}{c}{17.52} & \multicolumn{1}{c}{17.85} & \multicolumn{1}{c}{18.23} & \multicolumn{1}{c}{18.19} & \multicolumn{1}{c}{17.42} &\multicolumn{1}{c}{18.03} & \multicolumn{1}{c}{18.14}& \multicolumn{1}{c}{18.26} & \multicolumn{1}{c}{17.58} & \multicolumn{1}{c}{15.67} & \multicolumn{1}{c}{\underline{12.54}} & \multicolumn{1}{c}{\textbf{12.43}}\\

    & \multirow {1}{*}{INS}   &  \multicolumn{1}{c}{45.65} & \multicolumn{1}{c}{45.19} & \multicolumn{1}{c}{43.86} & \multicolumn{1}{c}{49.18} & \multicolumn{1}{c}{37.22} 
    &\multicolumn{1}{c}{32.99} & \multicolumn{1}{c}{31.02}& \multicolumn{1}{c}{42.01} & \multicolumn{1}{c}{44.14} & \multicolumn{1}{c}{50.36} & \multicolumn{1}{c}{\underline{51.31}}& \multicolumn{1}{c}{\textbf{51.54}}\\

         & \multirow {1}{*} {DEL} & \multicolumn{1}{c}{13.43} & \multicolumn{1}{c}{14.17} & \multicolumn{1}{c}{15.18} & \multicolumn{1}{c}{14.73} & \multicolumn{1}{c}{12.36} 
         &\multicolumn{1}{c}{13.08} & \multicolumn{1}{c}{13.29}& \multicolumn{1}{c}{14.21} & \multicolumn{1}{c}{13.64} & \multicolumn{1}{c}{11.68} & \multicolumn{1}{c}{\underline{10.21}}& \multicolumn{1}{c}{\textbf{9.96}}\\

         & \multirow {1}{*}{ADP}   & \multicolumn{1}{c}{22.46} & \multicolumn{1}{c}{22.35} & \multicolumn{1}{c}{29.13} & \multicolumn{1}{c}{24.38} & \multicolumn{1}{c}{36.98} 
         &\multicolumn{1}{c}{35.79} & \multicolumn{1}{c}{41.73}& \multicolumn{1}{c}{30.75} & \multicolumn{1}{c}{37.62} & \multicolumn{1}{c}{16.73} & \multicolumn{1}{c}{\underline{15.49}}& \multicolumn{1}{c}{\textbf{15.31}}\\

        & \multirow {1}{*}{PIC}   &    \multicolumn{1}{c}{23.16} & \multicolumn{1}{c}{24.42} & \multicolumn{1}{c}{22.34} & \multicolumn{1}{c}{24.59} & \multicolumn{1}{c}{17.65} 
        &\multicolumn{1}{c}{13.12} & \multicolumn{1}{c}{20.69}& \multicolumn{1}{c}{22.13} & \multicolumn{1}{c}{22.17} & \multicolumn{1}{c}{27.11} & \multicolumn{1}{c}{\underline{30.28}}& \multicolumn{1}{c}{\textbf{31.40}}\\

        & \multirow {1}{*}{SIC}   &  \multicolumn{1}{c}{65.93} & \multicolumn{1}{c}{67.94} & \multicolumn{1}{c}{54.75} & \multicolumn{1}{c}{63.95} & \multicolumn{1}{c}{53.36} & \multicolumn{1}{c}{58.35} & \multicolumn{1}{c}{57.27} & \multicolumn{1}{c}{62.84} & \multicolumn{1}{c}{69.11} & \multicolumn{1}{c}{69.63} & \multicolumn{1}{c}{\underline{71.39}}& \multicolumn{1}{c}{\textbf{71.97}}\\

        & \multirow {1}{*}{AIC}   &  \multicolumn{1}{c}{75.64} & \multicolumn{1}{c}{75.52} & \multicolumn{1}{c}{57.06} & \multicolumn{1}{c}{71.53} & \multicolumn{1}{c}{51.68} & \multicolumn{1}{c}{55.82} & \multicolumn{1}{c}{53.82} & \multicolumn{1}{c}{67.15} & \multicolumn{1}{c}{75.41} & \multicolumn{1}{c}{77.89} & \multicolumn{1}{c}{\underline{80.03}}& \multicolumn{1}{c}{\textbf{80.24}}\\

\midrule
    \multirow{8}{*}{DN}

    & \multirow {1}{*}{NEG}   & \multicolumn{1}{c}{57.40} & \multicolumn{1}{c}{57.16} & \multicolumn{1}{c}{58.01} & \multicolumn{1}{c}{56.63} & \multicolumn{1}{c}{40.74} 
    &\multicolumn{1}{c}{37.31} & \multicolumn{1}{c}{36.67}& \multicolumn{1}{c}{56.79} & \multicolumn{1}{c}{56.96} & \multicolumn{1}{c}{57.32} & \multicolumn{1}{c}{\underline{58.94}}& \multicolumn{1}{c}{\textbf{59.17}}\\

    & \multirow {1}{*}{POS}   &  \multicolumn{1}{c}{17.75} & \multicolumn{1}{c}{17.81} & \multicolumn{1}{c}{18.87} & \multicolumn{1}{c}{18.67} & \multicolumn{1}{c}{17.31} 
    &\multicolumn{1}{c}{17.46} & \multicolumn{1}{c}{17.38}& \multicolumn{1}{c}{17.84} & \multicolumn{1}{c}{17.62} & \multicolumn{1}{c}{16.82} & \multicolumn{1}{c}{\underline{13.65}} & \multicolumn{1}{c}{\textbf{13.14}}\\

    & \multirow {1}{*}{INS}   & \multicolumn{1}{c}{51.09} & \multicolumn{1}{c}{50.89} & \multicolumn{1}{c}{50.63} & \multicolumn{1}{c}{50.41} & \multicolumn{1}{c}{37.58} 
    &\multicolumn{1}{c}{33.31} & \multicolumn{1}{c}{31.32}& \multicolumn{1}{c}{50.44} & \multicolumn{1}{c}{50.60} & \multicolumn{1}{c}{50.98} & \multicolumn{1}{c}{\underline{51.63}} & \multicolumn{1}{c}{\textbf{51.97}}\\

         & \multirow {1}{*}{DEL}   & \multicolumn{1}{c}{13.61} & \multicolumn{1}{c}{13.63} & \multicolumn{1}{c}{13.29} & \multicolumn{1}{c}{15.31} & \multicolumn{1}{c}{13.26} 
         &\multicolumn{1}{c}{13.27} & \multicolumn{1}{c}{13.54}& \multicolumn{1}{c}{14.34} & \multicolumn{1}{c}{13.85} & \multicolumn{1}{c}{13.02} & \multicolumn{1}{c}{\underline{10.28}}& \multicolumn{1}{c}{\textbf{10.04}}\\

         & \multirow {1}{*}{ADP}   & \multicolumn{1}{c}{17.46} & \multicolumn{1}{c}{17.01} & \multicolumn{1}{c}{19.45} & \multicolumn{1}{c}{17.13} & \multicolumn{1}{c}{35.61}
         &\multicolumn{1}{c}{34.51} & \multicolumn{1}{c}{40.04}& \multicolumn{1}{c}{20.21} & \multicolumn{1}{c}{24.23} & \multicolumn{1}{c}{13.42} & \multicolumn{1}{c}{\underline{12.06}}& \multicolumn{1}{c}{\textbf{11.98}}\\

        & \multirow {1}{*}{PIC}   &  \multicolumn{1}{c}{34.68} & \multicolumn{1}{c}{35.21} & \multicolumn{1}{c}{34.13} & \multicolumn{1}{c}{31.22} & \multicolumn{1}{c}{22.35} 
        &\multicolumn{1}{c}{16.62} & \multicolumn{1}{c}{26.18}& \multicolumn{1}{c}{31.05} & \multicolumn{1}{c}{33.81} & \multicolumn{1}{c}{39.54} & \multicolumn{1}{c}{\underline{46.53}}& \multicolumn{1}{c}{\textbf{47.28}}\\

        & \multirow {1}{*}{SIC}   &  \multicolumn{1}{c}{75.62} & \multicolumn{1}{c}{74.75} & \multicolumn{1}{c}{74.72} & \multicolumn{1}{c}{73.94} & \multicolumn{1}{c}{54.59} & \multicolumn{1}{c}{58.55} & \multicolumn{1}{c}{57.66} & \multicolumn{1}{c}{72.93} & \multicolumn{1}{c}{74.34} & \multicolumn{1}{c}{77.71} & \multicolumn{1}{c}{\underline{80.96}}& \multicolumn{1}{c}{\textbf{81.49}}\\

        & \multirow {1}{*}{AIC}   &  \multicolumn{1}{c}{74.22} & \multicolumn{1}{c}{71.82} & \multicolumn{1}{c}{72.65} & \multicolumn{1}{c}{70.21} & \multicolumn{1}{c}{54.74} & \multicolumn{1}{c}{54.56} & \multicolumn{1}{c}{56.08} & \multicolumn{1}{c}{70.63} & \multicolumn{1}{c}{71.82} & \multicolumn{1}{c}{75.22} & \multicolumn{1}{c}{\underline{79.01}}& \multicolumn{1}{c}{\textbf{79.35}}\\

        \bottomrule
      \end{tabular}}

  \end{center}
    \caption{Results on the IN dataset (CN and DN models): For POS, DEL and ADP, lower is better. For NEG, INS, PIC, SIC and AIC, higher is better.}
  \label{tab:cnn_backbones_metrics_exp_appendix}
    \end{table*}

%% file: latex/tables/vit_explanation_appendix.tex
\begin{table}

  \begin{center}

  \scalebox{0.9}{
    \begin{tabular}{@{}lc@{}lc@{}lc@{}lc@{}}
    \toprule
      & & \multicolumn{1}{l}{T-Attr} & \multicolumn{1}{l}{GAE} & \multicolumn{1}{l}{pBEE}& \multicolumn{1}{l}{fBEE} \\
    \midrule

    \multirow{8}{*}{ViT-S}
    
        & \multirow {1}{*}{NEG} & \multicolumn{1}{l}{53.29} & \multicolumn{1}{l}{52.81} &  \multicolumn{1}{l}{\underline{57.74}} & \multicolumn{1}{l}{\textbf{58.07}}\\

        & \multirow {1}{*}{POS} & \multicolumn{1}{l}{14.16} & \multicolumn{1}{l}{14.75} &  \multicolumn{1}{l}{\underline{11.84}}& \multicolumn{1}{l}{\textbf{11.62}}\\

        & \multirow {1}{*}{INS} & \multicolumn{1}{l}{45.72} & \multicolumn{1}{l}{45.21} & \multicolumn{1}{l}{\underline{48.66}}& \multicolumn{1}{l}{\textbf{49.33}} \\

         & \multirow {1}{*}{DEL} & \multicolumn{1}{l}{11.28} & \multicolumn{1}{l}{11.92} &  \multicolumn{1}{l}{\underline{9.56}}& \multicolumn{1}{l}{\textbf{9.42}} \\

         & \multirow {1}{*}{ADP} & \multicolumn{1}{l}{51.94} & \multicolumn{1}{l}{36.98} &  \multicolumn{1}{l}{\underline{34.90}} & \multicolumn{1}{l}{\textbf{34.74}}\\

        & \multirow {1}{*}{PIC} &   \multicolumn{1}{l}{13.67} &  \multicolumn{1}{l}{8.68}&  \multicolumn{1}{l}{\underline{22.48}}& \multicolumn{1}{l}{\textbf{24.63}}\\

        & \multirow {1}{*}{SIC} &  \multicolumn{1}{l}{69.46} & \multicolumn{1}{l}{70.19} & \multicolumn{1}{l}{\underline{72.85}}& \multicolumn{1}{l}{\textbf{73.16}}\\

        & \multirow {1}{*}{AIC} &  \multicolumn{1}{l}{63.86} & \multicolumn{1}{l}{64.49}  &  \multicolumn{1}{l}{\underline{68.29}}& \multicolumn{1}{l}{\textbf{68.44}}\\
            
        \bottomrule
  \end{tabular}}
  
  \end{center}

\caption{Results on the IN dataset (ViT-S model): For POS, DEL and ADP, lower is better. For NEG, INS, PIC, SIC and AIC, higher is better.
  }
  \label{tab:vit_backbones_metrics_exp_appendix}
\end{table}

%% file: latex/tables/ablation_study.tex
\begin{table*}[ht!]
  \begin{center}
    
  \scalebox{1}{
    \begin{tabular}{@{}lc@{}lc@{}lc@{}lc@{}lc@{}lc@{}}%
    \toprule
      & & \multicolumn{1}{l}{NEG} & \multicolumn{1}{l}{POS} & \multicolumn{1}{l}{INS} & \multicolumn{1}{l}{DEL} & \multicolumn{1}{l}{ADP} & \multicolumn{1}{l}{PIC} & \multicolumn{1}{l}{SIC} & \multicolumn{1}{l}{AIC}\\
    \midrule
        & \multirow{1}{*}{IG-fBEE}  &  \multicolumn{1}{l}{51.26}& \multicolumn{1}{l}{16.10}  & \multicolumn{1}{l}{42.13} & \multicolumn{1}{l}{13.01} &  \multicolumn{1}{l}{21.43}& \multicolumn{1}{l}{28.59}  & \multicolumn{1}{l}{70.62} & \multicolumn{1}{l}{68.84}\\



        & \multirow{1}{*}{ACT-IG}  &  \multicolumn{1}{l}{55.41}& \multicolumn{1}{l}{17.39} & \multicolumn{1}{l}{47.53} & \multicolumn{1}{l}{13.72} &  \multicolumn{1}{l}{17.21}& \multicolumn{1}{l}{36.54}  & \multicolumn{1}{l}{76.85} & \multicolumn{1}{l}{75.48}\\

        & \multirow{1}{*}{ncBEE}  &  \multicolumn{1}{l}{58.43}& \multicolumn{1}{l}{14.25}  & \multicolumn{1}{l}{50.19} & \multicolumn{1}{l}{11.56} &  \multicolumn{1}{l}{13.35}& \multicolumn{1}{l}{48.29}  & \multicolumn{1}{l}{80.94} & \multicolumn{1}{l}{78.66}\\

        & \multirow{1}{*}{pBEE}  &  \multicolumn{1}{l}{59.10}& \multicolumn{1}{l}{13.69}  & \multicolumn{1}{l}{51.15} & \multicolumn{1}{l}{11.19} &  \multicolumn{1}{l}{11.35}& \multicolumn{1}{l}{48.22}  & \multicolumn{1}{l}{81.23} & \multicolumn{1}{l}{78.45}\\

        & \multirow{1}{*}{fBEE}  &  \multicolumn{1}{l}{59.38}& \multicolumn{1}{l}{13.47}  & \multicolumn{1}{l}{51.73} & \multicolumn{1}{l}{10.42} &  \multicolumn{1}{l}{11.09}& \multicolumn{1}{l}{48.86}  & \multicolumn{1}{l}{81.51} & \multicolumn{1}{l}{79.21}\\

        \midrule
  \end{tabular}}
  \end{center}
\caption{Ablation Study: Comparing three alternatives for configurations of BEE.
  }
  \label{tab:ablation_study_table}
\end{table*}

%% file: latex/tables/ablation_study_bee_pi_smooth.tex
\begin{table*}[ht!]
  \begin{center}

  \scalebox{0.8}{
    \begin{tabular}{@{}lc@{}lc@{}lc@{}lc@{}lc@{}lc@{}}%
    \toprule
      & & \multicolumn{1}{l}{NEG} & \multicolumn{1}{l}{POS} & \multicolumn{1}{l}{INS} & \multicolumn{1}{l}{DEL} & \multicolumn{1}{l}{ADP} & \multicolumn{1}{l}{PIC} & \multicolumn{1}{l}{SIC} & \multicolumn{1}{l}{AIC}\\
    \midrule

        & \multirow{1}{*}{IG}  &  \multicolumn{1}{l}{45.66}& \multicolumn{1}{l}{17.24}  & \multicolumn{1}{l}{39.87} & \multicolumn{1}{l}{13.49} &  \multicolumn{1}{l}{37.52}& \multicolumn{1}{l}{19.94}  & \multicolumn{1}{l}{54.67} & \multicolumn{1}{l}{51.92}\\

        & \multirow{1}{*}{BIG}  &  \multicolumn{1}{l}{42.25}& \multicolumn{1}{l}{17.44}  & \multicolumn{1}{l}{36.04} & \multicolumn{1}{l}{13.95} &  \multicolumn{1}{l}{40.85}& \multicolumn{1}{l}{24.53}  & \multicolumn{1}{l}{56.98} & \multicolumn{1}{l}{53.36}\\

        & \multirow{1}{*}{GIG}  &  \multicolumn{1}{l}{43.97}& \multicolumn{1}{l}{17.68}  & \multicolumn{1}{l}{37.92} & \multicolumn{1}{l}{14.18} &  \multicolumn{1}{l}{35.28}& \multicolumn{1}{l}{18.72}  & \multicolumn{1}{l}{55.04} & \multicolumn{1}{l}{53.38}\\

        & \multirow{1}{*}{IG-fBEE}  &  \multicolumn{1}{l}{51.26}& \multicolumn{1}{l}{16.10}  & \multicolumn{1}{l}{42.13} & \multicolumn{1}{l}{13.01} &  \multicolumn{1}{l}{21.43}& \multicolumn{1}{l}{28.59}  & \multicolumn{1}{l}{70.62} & \multicolumn{1}{l}{68.84}\\

        & \multirow{1}{*}{BIG-fBEE}  &  \multicolumn{1}{l}{51.12}& \multicolumn{1}{l}{15.94}  & \multicolumn{1}{l}{41.93} & \multicolumn{1}{l}{12.96} &  \multicolumn{1}{l}{21.19}& \multicolumn{1}{l}{29.13}  & \multicolumn{1}{l}{72.19} & \multicolumn{1}{l}{69.38}\\

        & \multirow{1}{*}{GIG-fBEE}  &  \multicolumn{1}{l}{51.82}& \multicolumn{1}{l}{15.83}  & \multicolumn{1}{l}{42.45} & \multicolumn{1}{l}{12.81} &  \multicolumn{1}{l}{20.48}& \multicolumn{1}{l}{29.90}  & \multicolumn{1}{l}{75.32} & \multicolumn{1}{l}{74.61}\\

        & \multirow{1}{*}{pBEE}  &  \multicolumn{1}{l}{59.10}& \multicolumn{1}{l}{13.69}  & \multicolumn{1}{l}{51.15} & \multicolumn{1}{l}{11.19} &  \multicolumn{1}{l}{11.35}& \multicolumn{1}{l}{48.22}  & \multicolumn{1}{l}{81.23} & \multicolumn{1}{l}{78.45}\\

        & \multirow{1}{*}{fBEE}  &  \multicolumn{1}{l}{59.38}& \multicolumn{1}{l}{13.47}  & \multicolumn{1}{l}{51.73} & \multicolumn{1}{l}{10.42} &  \multicolumn{1}{l}{11.09}& \multicolumn{1}{l}{48.86}  & \multicolumn{1}{l}{81.51} & \multicolumn{1}{l}{79.21}\\




        \midrule
  \end{tabular}}
  \end{center}
    \caption{Ablation Study: The benefit of applying BEE for different path integration methods.
  }
  \label{tab:ablation_study_pi_table}
\end{table*}

%% file: latex/tables/ablation_samples.tex
\begin{table*}[ht!]
  \begin{center}

  \scalebox{0.8}{
    \begin{tabular}{@{}lc@{}lc@{}lc@{}lc@{}lc@{}lc@{}}%
    \toprule
      & & \multicolumn{1}{l}{NEG} & \multicolumn{1}{l}{POS} & \multicolumn{1}{l}{INS} & \multicolumn{1}{l}{DEL} & \multicolumn{1}{l}{ADP} & \multicolumn{1}{l}{PIC} & \multicolumn{1}{l}{SIC} & \multicolumn{1}{l}{AIC}\\
    \midrule
        \multirow{5}{*}{fBEE}
        & \multirow{1}{*}{T=4}  &  \multicolumn{1}{l}{58.46}& \multicolumn{1}{l}{15.13}  & \multicolumn{1}{l}{50.94} & \multicolumn{1}{l}{11.98} &  \multicolumn{1}{l}{13.10}& \multicolumn{1}{l}{45.16}  & \multicolumn{1}{l}{77.89} & \multicolumn{1}{l}{78.02}\\

        & \multirow{1}{*}{T=8}  &  \multicolumn{1}{l}{59.38}& \multicolumn{1}{l}{13.47} & \multicolumn{1}{l}{51.73} & \multicolumn{1}{l}{10.42} &  \multicolumn{1}{l}{11.09}& \multicolumn{1}{l}{48.86}  & \multicolumn{1}{l}{81.51} & \multicolumn{1}{l}{79.21}\\

        & \multirow{1}{*}{T=16}  &  \multicolumn{1}{l}{60.44}& \multicolumn{1}{l}{12.39}  & \multicolumn{1}{l}{52.04} & \multicolumn{1}{l}{9.85} &  \multicolumn{1}{l}{10.48}& \multicolumn{1}{l}{50.71}  & \multicolumn{1}{l}{81.72} & \multicolumn{1}{l}{79.95}\\

        & \multirow{1}{*}{T=32}  &  \multicolumn{1}{l}{61.59}& \multicolumn{1}{l}{11.41}  & \multicolumn{1}{l}{52.89} & \multicolumn{1}{l}{9.03} &  \multicolumn{1}{l}{9.36}& \multicolumn{1}{l}{53.87}  & \multicolumn{1}{l}{82.16} & \multicolumn{1}{l}{81.48}\\

        & \multirow{1}{*}{T=64}  &  \multicolumn{1}{l}{62.49}& \multicolumn{1}{l}{10.85}  & \multicolumn{1}{l}{53.75} & \multicolumn{1}{l}{8.01} &  \multicolumn{1}{l}{7.98}& \multicolumn{1}{l}{56.14}  & \multicolumn{1}{l}{82.69} & \multicolumn{1}{l}{82.89}\\
        \midrule

    \multirow{5}{*}{pBEE}
        & \multirow{1}{*}{T=4}  &  \multicolumn{1}{l}{58.39}& \multicolumn{1}{l}{15.42}  & \multicolumn{1}{l}{50.66} & \multicolumn{1}{l}{12.12} &  \multicolumn{1}{l}{13.24}& \multicolumn{1}{l}{44.81}  & \multicolumn{1}{l}{77.72} & \multicolumn{1}{l}{77.84}\\

        & \multirow{1}{*}{T=8}  &  \multicolumn{1}{l}{59.10}& \multicolumn{1}{l}{13.69} & \multicolumn{1}{l}{51.15} & \multicolumn{1}{l}{11.19} &  \multicolumn{1}{l}{11.35}& \multicolumn{1}{l}{48.22}  & \multicolumn{1}{l}{81.23} & \multicolumn{1}{l}{78.45}\\

        & \multirow{1}{*}{T=16}  &  \multicolumn{1}{l}{59.65}& \multicolumn{1}{l}{12.84}  & \multicolumn{1}{l}{51.79} & \multicolumn{1}{l}{10.36} &  \multicolumn{1}{l}{11.03}& \multicolumn{1}{l}{49.85}  & \multicolumn{1}{l}{81.31} & \multicolumn{1}{l}{79.93}\\

        & \multirow{1}{*}{T=32}  &  \multicolumn{1}{l}{60.91}& \multicolumn{1}{l}{12.09}  & \multicolumn{1}{l}{52.63} & \multicolumn{1}{l}{9.81} &  \multicolumn{1}{l}{9.68}& \multicolumn{1}{l}{51.13}  & \multicolumn{1}{l}{81.78} & \multicolumn{1}{l}{81.24}\\

        & \multirow{1}{*}{T=64}  &  \multicolumn{1}{l}{61.57}& \multicolumn{1}{l}{11.43}  & \multicolumn{1}{l}{53.12} & \multicolumn{1}{l}{8.47} &  \multicolumn{1}{l}{8.42}& \multicolumn{1}{l}{53.92}  & \multicolumn{1}{l}{82.14} & \multicolumn{1}{l}{82.17}\\
        \midrule
  \end{tabular}}
  \end{center}
    \caption{Ablation Study on the parameter $T$ - the number of sampled baselines per example.}
  \label{tab:samples_ablation}
\end{table*}

%% file: latex/tables/ablation_of_steps.tex
\begin{table*}[ht!]
  \begin{center}

  \scalebox{0.8}{
    \begin{tabular}{@{}lc@{}lc@{}lc@{}lc@{}lc@{}lc@{}}%
    \toprule
      & & \multicolumn{1}{l}{NEG} & \multicolumn{1}{l}{POS} & \multicolumn{1}{l}{INS} & \multicolumn{1}{l}{DEL} & \multicolumn{1}{l}{ADP} & \multicolumn{1}{l}{PIC} & \multicolumn{1}{l}{SIC} & \multicolumn{1}{l}{AIC} & \multicolumn{1}{l}{\textbf{Runtime (ms)}}\\
    \midrule

        & \multirow{1}{*}{n=5}  &  \multicolumn{1}{l}{54.86}& \multicolumn{1}{l}{18.16} & \multicolumn{1}{l}{47.05} & \multicolumn{1}{l}{13.91} &  \multicolumn{1}{l}{17.83}& \multicolumn{1}{l}{35.12}  & \multicolumn{1}{l}{76.29} & \multicolumn{1}{l}{75.10} & \multicolumn{1}{c}{198}\\
        
        & \multirow{1}{*}{n=10}  &  \multicolumn{1}{l}{55.41}& \multicolumn{1}{l}{17.39} & \multicolumn{1}{l}{47.53} & \multicolumn{1}{l}{13.72} &  \multicolumn{1}{l}{17.21}& \multicolumn{1}{l}{36.54}  & \multicolumn{1}{l}{76.85} & \multicolumn{1}{l}{75.48} & \multicolumn{1}{c}{209}\\

        & \multirow{1}{*}{n=50}  &  \multicolumn{1}{l}{55.49}& \multicolumn{1}{l}{17.31} & \multicolumn{1}{l}{47.58} & \multicolumn{1}{l}{13.67} &  \multicolumn{1}{l}{17.30}& \multicolumn{1}{l}{36.46}  & \multicolumn{1}{l}{76.78} & \multicolumn{1}{l}{75.43} & \multicolumn{1}{c}{321}\\

        & \multirow{1}{*}{n=100}  &  \multicolumn{1}{l}{55.52}& \multicolumn{1}{l}{17.28} & \multicolumn{1}{l}{47.61} & \multicolumn{1}{l}{13.65} &  \multicolumn{1}{l}{17.43}& \multicolumn{1}{l}{36.31}  & \multicolumn{1}{l}{76.74} & \multicolumn{1}{l}{75.40} & \multicolumn{1}{c}{496}\\

        \midrule
  \end{tabular}}
  \end{center}
    \caption{Ablation Study: Comparison between different number of interpolation steps.
  }
  \label{tab:steps_ablation}
\end{table*}


        


